%% file: main.tex
\documentclass[sigconf, nonacm]{acmart}

\newcommand\vldbdoi{10.14778/3489496.3489508}
\newcommand\vldbpages{272 - 284}
\newcommand\vldbvolume{15}
\newcommand\vldbissue{2}
\newcommand\vldbyear{2022}
\newcommand\vldbauthors{\authors}
\newcommand\vldbtitle{\shorttitle} 
\newcommand\vldbavailabilityurl{http://vldb.org/pvldb/format_vol14.html}
\newcommand\vldbpagestyle{empty}

\newcommand\revise[1]{{\textcolor{black}{#1}}\xspace}
\definecolor{revision}{RGB}{255,255,255}
\newcommand{\revision}[1]{{\color{revision} #1}\xspace}
\newcommand{\revisionstart}{\begin{color}{revision}}
\newcommand{\revisionend}{~\!\!\end{color}}

\newtheorem{manualtheoreminner}{Theorem}
\newenvironment{manualtheorem}[1]{%
  \renewcommand\themanualtheoreminner{#1}%
  \manualtheoreminner
}{\endmanualtheoreminner}

\newtheorem{manualpropositioninner}{Proposition}
\newenvironment{manualproposition}[1]{%
  \renewcommand\themanualpropositioninner{#1}%
  \manualpropositioninner
}{\endmanualtheoreminner}

\def\TR{1} 

\input{paper-command}

\begin{document}
\title{Learning to be a Statistician: Learned Estimator for Number of Distinct Values}

\author{Renzhi Wu}
\authornote{Work done at Alibaba Group.}
\affiliation{%
 \institution{Georgia Institute of Technology}
}
\email{renzhiwu@gatech.edu}

\author{Bolin Ding}
\affiliation{%
 \institution{Alibaba Group}
}
\email{bolin.ding@alibaba-inc.com}

\author{Xu Chu}
\affiliation{%
 \institution{Georgia Institute of Technology}
}
\email{xu.chu@cc.gatech.edu}

\author{Zhewei Wei}
\affiliation{
 \institution{Renmin University of China}
}
\email{zhewei@ruc.edu.cn}

\author{Xiening Dai}
\affiliation{%
 \institution{Alibaba Group}
}
\email{xiening.dai@alibaba-inc.com}

\author{Tao Guan, Jingren Zhou}
\affiliation{%
 \institution{Alibaba Group}
}
\email{{tony.guan, jingren.zhou}@alibaba-inc.com}
%
%
%
\renewcommand{\shortauthors}{R. Wu, B. Ding, X. Chu, Z. Wei, X. Dai, T. Guan, J. Zhou}
\renewcommand{\authors}{
Renzhi Wu,
Bolin Ding,
Xu Chu,
Zhewei Wei,
Xiening Dai,
Tao Guan,
Jingren Zhou}
\begin{abstract}
Estimating the number of distinct values (NDV) in a column is useful for many tasks in database systems, such as columnstore compression and data profiling. In this work, we focus on how to derive accurate NDV estimations from random (online/offline) samples. Such efficient estimation is critical for tasks where it is prohibitive to scan the data even once.
%
Existing sample-based estimators typically rely on heuristics or assumptions and do not have robust performance across different datasets as the assumptions on data can easily break.
On the other hand, deriving an estimator from a principled formulation such as maximum likelihood estimation is very challenging due to the complex structure of the formulation.
We propose to formulate the NDV estimation task in a supervised learning framework, and aim to learn a model as the estimator. To this end, we need to answer several questions: i) how to make the learned model workload agnostic; ii) how to obtain training data; iii) how to perform model training.
We derive conditions of the learning framework under which the learned model is {\em workload agnostic}, in the sense that 
the model/estimator can be trained with synthetically generated training data, and then deployed into any data warehouse simply as, \eg, user-defined functions (UDFs), to offer efficient (within microseconds \revise{on CPU}) and accurate NDV estimations for {\em unseen tables and workloads}. 
We compare the learned estimator with the state-of-the-art sample-based estimators on nine real-world datasets to demonstrate its superior estimation accuracy.
We publish our \revise{code for training data generation, model training, and the} learned estimator online for reproducibility.
\end{abstract}

\maketitle

\pagestyle{\vldbpagestyle}
\begingroup\small\noindent\raggedright\textbf{PVLDB Reference Format:}\\
\vldbauthors. \vldbtitle. PVLDB, \vldbvolume(\vldbissue): \vldbpages, \vldbyear.\\
\href{https://doi.org/\vldbdoi}{doi:\vldbdoi}
\endgroup
\begingroup
\renewcommand\thefootnote{}\footnote{\noindent
This work is licensed under the Creative Commons BY-NC-ND 4.0 International License. Visit \url{https://creativecommons.org/licenses/by-nc-nd/4.0/} to view a copy of this license. For any use beyond those covered by this license, obtain permission by emailing \href{mailto:info@vldb.org}{info@vldb.org}. Copyright is held by the owner/author(s). Publication rights licensed to the VLDB Endowment. \\
\raggedright Proceedings of the VLDB Endowment, Vol. \vldbvolume, No. \vldbissue\ %
ISSN 2150-8097. \\
\href{https://doi.org/\vldbdoi}{doi:\vldbdoi} \\
}\addtocounter{footnote}{-1}\endgroup


\input{paper-intro}

\input{paper-pre}

\input{paper-overview}

\input{paper-method}
\input{paper-experiments}
\section{Conclusion and Future work}
In this paper, we consider a fundamental question: \textit{whether it is possible to train a workload-agnostic machine learning model to approximate principled statistical estimators such as maximum likelihood estimators (MLE)}. We provide a positive answer to this question on the concrete task of estimating the number of distinct values (NDV) of a population from a small sample.
We formulate the sample-based NDV estimation problem as an MLE problem which, however, is difficult to be solved even approximately. We propose a learning-to-estimate framework to train a workload-agnostic model to approximate the MLE estimator.
%
%
%
%
Extensive experiments on nine datasets from diverse domains demonstrate that our learned estimator is robust and outperforms all baselines significantly. 

For future work, we would like to extend our learning-to-estimate method to learn estimators for other properties whose MLE is difficult to obtain, \eg, entropy and distance to uniformity.


\if\TR 1
\input{paper-appendix}
\fi
\bibliographystyle{ACM-Reference-Format}
\bibliography{main}

\end{document}

%% file: paper-command.tex
\usepackage{amsmath, amsfonts}
\usepackage{mathtools, nccmath}
\usepackage{multirow}
\usepackage[linesnumbered,ruled]{algorithm2e}
\usepackage{enumitem}
\usepackage{makecell}
\usepackage{adjustbox}
\usepackage{xspace}

\DeclarePairedDelimiter{\nint}\lfloor\rceil
\DeclareMathOperator*{\argmax}{arg\,max}

\newtheorem{lemma}{Lemma}

\newtheorem{proposition}[lemma]{Proposition}

\newcommand{\stitle}[1]{\noindent{\bf #1}}
\newcommand{\sstitle}[1]{\noindent$\bullet$~{\em #1}}

\renewcommand{\qed}{\hfill $\framebox(6,6){}$}

\newcommand{\ctab}{Table~}

\newcommand{\csec}{Section~}

\newcommand{\cthm}{Theorem~}

\newcommand{\cprop}{Proposition~}

\newcommand{\cfig}{Figure~}
\newcommand{\cfigs}{Figures~}

\newcommand{\ie}{{\em i.e.}\xspace}
\newcommand{\eg}{{\em e.g.}\xspace}

\newcommand{\etal}{{\em et al.}\xspace}

\newcommand{\squishlist}{
	\begin{list}{$\bullet$}{
		\setlength{\itemsep}{0pt}
		\setlength{\parsep}{3pt}
		\setlength{\topsep}{3pt}
		\setlength{\partopsep}{0pt}
		\setlength{\leftmargin}{1.0em}
		\setlength{\labelwidth}{1em}
		\setlength{\labelsep}{0.5em}
   }
}

\newcommand{\squishenum}{
	
	\begin{list}{\usecounter{scount}}{
		\setlength{\itemsep}{0pt}
		\setlength{\parsep}{3pt}
		\setlength{\topsep}{3pt}
		\setlength{\partopsep}{0pt}
		\setlength{\leftmargin}{1.2em}
		\setlength{\labelwidth}{1em}
		\setlength{\labelsep}{0.5em}
	}
}

\newcommand{\squishend}{
	\end{list}
}

\newcommand{\eat}[1]{}

\newcommand{\pr}[1]{\mathbb{P}\!\left(#1\right)\xspace}
\newcommand{\prp}[1]{\mathbb{P}_{\!\!\traindist}\!\left(#1\right)\xspace}

\newcommand{\epinline}[1]{{\bf E}[#1]\xspace}

\newcommand{\bigohinline}[1]{{\rm O}(#1)\xspace}

\newcommand{\bigomegainline}[1]{{\rm \Omega}(#1)\xspace}

\newcommand{\col}{\mathbf{C}\xspace}
\newcommand{\sample}{\mathbf{S}\xspace}
\newcommand{\traindist}{\mathcal{A}\xspace}
\newcommand{\trainprof}{\mathcal{F}\xspace}
\newcommand{\trainprofp}{\mathcal{F'}\xspace}
\newcommand{\dom}{\Omega\xspace}

\newcommand{\est}{{\sf h}\xspace}

\newcommand{\estclass}{\mathcal{H}\xspace}
\newcommand{\err}{{\rm error}\xspace}

%% file: paper-intro.tex
\section{Introduction}
Estimating number of distinct values (NDV), also known as cardinality estimation, is a fundamental problem with numerous applications~\cite{harmouch2017cardinality,haas1995sampling,nath2008synopsis,mohamadi2017ntcard}. It has been extensively studied in many research communities including databases~\cite{haas1995sampling,charikar2000towards}, networks~\cite{nath2008synopsis, cohen2019cardinality}, bioinformatics~\cite{mohamadi2017ntcard}, and statistics~\cite{bunge1993estimating,haas1998estimating, pavlichin2019approximate}. 

The methods for estimating NDV in the absence of an index can be classified into two categories: {\em sketch based methods} and {\em sampling based methods}~\cite{metwally2008go}. 
Sketch based methods scan the entire dataset once, followed by sorting/hashing rows, and create a sketch that is used to estimate NDV~\cite{harmouch2017cardinality}. 
Sampling based methods estimate NDV using statistics from a small sample, without needing to scan the entire dataset.
Generally, sketch based methods give more accurate estimation, but scanning and hashing the entire dataset can be prohibitively expensive in large data warehouses. Hashing techniques such as probabilistic counting help alleviate the memory requirements but still requires a full scan of the table.
When a full scan is not possible or the computation cost of a full scan is not affordable, sampling based methods are the only remaining alternatives which exam only a very small 
fraction of the table, \ie, a sample, and thus scale well with increasing data set.

In this paper, we focus on the problem of accurately estimating the number of distinct values from samples of large tables.
 
The first challenge is that, unlike some other statistical parameters, such as means and histograms, which can be accurately computed from small random samples, accurate NDV estimation from small samples has been proved to be an extremely difficult task (\eg, with theoretical lower bounds of error given in \cite{charikar2000towards}).

We can formulate sample-based NDV estimation using a principled method for estimating unknown parameters, the {\em maximum likelihood estimation} (MLE)~\cite{chambers2012maximum}. An MLE estimator is {\em workload-agnostic}: it is derived (analytically) before we see the real workloads. It solves an optimization problem, which maximizes the likelihood of observing a specific random sample, and gives an NDV estimation with desirable properties such as {\em consistency} and {\em efficiency}. However, the remaining steps are challenging (if not impossible): how to express the likelihood function and the optimization problem in a compact way, and how to (even approximately) solve it.

Due to the above challenges, this paper proposes and studies a more fundamental question: {\em whether it is possible to train a workload-agnostic machine learning model to approximate principled statistical estimators such as MLE estimators}, with the training set synthetically generated from a training distribution calibrated based on the properties of our estimation task, such that the learned model can be used on unseen workloads.


\begin{figure}
  \centering
  \includegraphics[width=\linewidth]{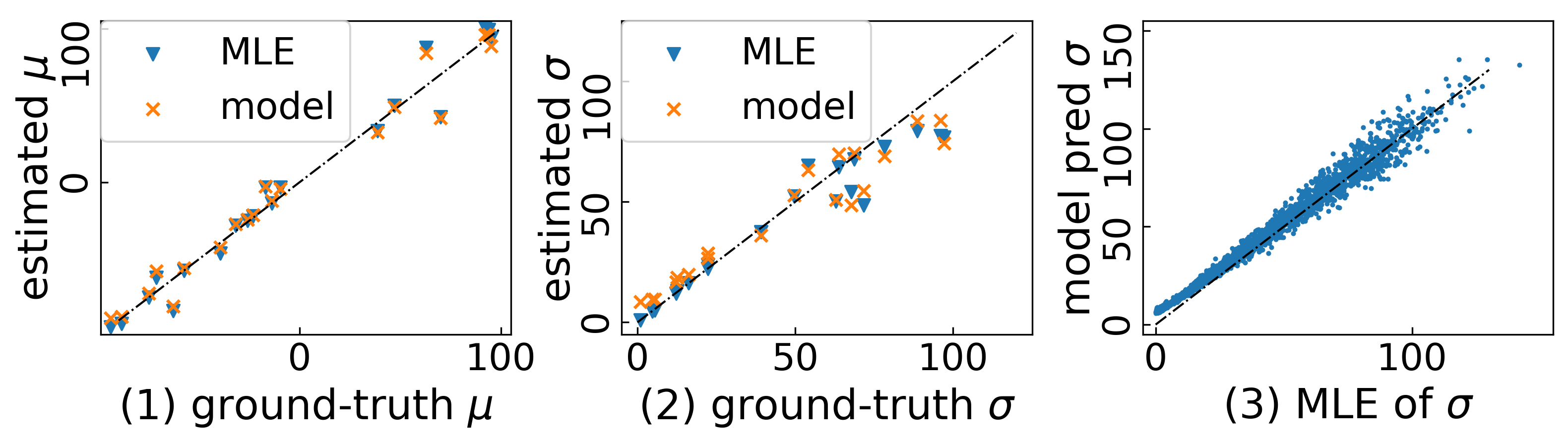}
  \caption{Evaluating trained models to estimate mean $\mu$ and STD $\sigma$ of Gaussian distributions: (1) Estimated $\mu$ via MLE and the trained model v.s. ground-truth $\mu$ on 20 data points. (2) Estimated $\sigma$ via MLE and the trained model v.s. ground-truth $\sigma$ on 20 data points. (3) Model predicted $\sigma$ v.s. MLE estimation of $\sigma$ in the closed form on 2000 data points.}
  \label{fig:illu_learn_to_est}
\end{figure}

\subsection*{Learning to be a Statistician: a Toy Example}
%
%
It is traditionally the statistician's job to derive a closed-form or numerical estimator as a function that takes an observed sample or sample statistics as the input for estimating unknown parameters of a statistical model.
On the other hand, it is known that any continuous function can be approximated by a neural network (with enough nodes and layers)~\cite{haykin1998nn}, which require three elements: a training set $A$, a class of model architectures $\estclass$, and a training algorithm. 
Thus, a natural question is whether we can train a neural network in a {\em workload-agnostic way} (without seeing any real dataset during training) to approximate, \eg, an MLE estimator. Following are two examples with positive answers to this question.




\stitle{Learning MLE of mean and STD.}
Let's consider the mean $\mu$ and standard deviation (STD) $\sigma$ of a Gaussian distribution. For a size-$k$ sample drawn from this distribution, $\sample=\{v_1,\dots,v_k\}$, the MLE estimations of $\mu$ and $\sigma$ are known to have closed forms: $\mu^{\sf MLE}=\sum_{i} v_i/k$ and $\sigma^{\sf MLE} = \sqrt{\sum_i(v_i-\mu^{\sf MLE})^2/k}$, respectively. 
Assume that we do not know the concrete form of $\mu^{\sf MLE}$ and $\sigma^{\sf MLE}$. How should we train machine learning models to approximate them?

We first need to prepare the training data. From a specific Gaussian distribution $\mathcal{N}(\mu,\sigma)$, we draw a size-$k$ sample $\sample$, and add {\em one} training data point $(\sample, \mu)$ to the training set $A_1$ for mean estimation, and {\em one} data point $(\sample, \sigma)$ to $A_2$ for STD estimation. Note that, here, each $\sample$ is a $k$-dim feature vector, and each $\mu$ and $\sigma$ are the (ground-truth) labels to be predicted for mean and STD, respectively. In the outer loop, we need to try different combinations of $\mu$ and $\sigma$ to add more training data points into $A_1$ and $A_2$. In this example, we generate $10^5$ pairs of $(\mu,\sigma)$, where each $\mu$ is randomly picked from a Gaussian distribution $\mathcal{N}(0,100)$ and each $\sigma$ is picked from a folded Gaussian distribution (\ie, $\sigma = |z|$ where $z \sim \mathcal{N}(0,100)$). Thus, we will have $10^5$ training data points in each of $A_1$ and $A_2$.



We train a regression model (multi-layer perception with one hidden layer of size $k$ and mean squared loss) using the training set $A_1$ for estimating $\mu$ and another model (with the same architecture) on $A_2$ for estimating $\sigma$, both with $\sample$ as feature.

We evaluate the trained estimator on a completely different test set. We generate pairs $(\mu, \sigma)$ with each $\mu\sim U(-100,100)$ and $\sigma\sim U(0,100)$ from uniform distributions. For each $(\mu, \sigma)$, we generate a test data point $(\sample, \mu)$ and a point $(\sample, \sigma)$ for a size-$k$ sample $\sample$ from $\mathcal{N}(\mu, \sigma)$. The results are reported in \cfig\ref{fig:illu_learn_to_est} (where we use $k=100$). \cfigs\ref{fig:illu_learn_to_est}(1)-(2) show that the estimations of mean/STD given by the trained model have very similar performance as those by the closed-form MLE estimations. The trained model and MLE even make the same mistakes for some test points with non-trivial errors. %
\cfig\ref{fig:illu_learn_to_est}(3) compares $\sigma^{\sf MLE}$ and the learned estimator on more data points for STD estimation, for which the MLE estimator has a more complex form than a linear model for the mean estimation.

\subsection*{Our Contributions and Solution Overview}
In this paper, we will formalize and investigate the learning framework illustrated in the above example, but for a more challenging task, {\em sample-based NDV estimation}. While the empirical evidence for mean/STD estimation sheds light on the possibility of approximating MLE estimators with trained models, there are several key questions to be resolved for general estimation tasks: i) how to generate training data and what features should be extracted for learning the model; ii) what model architecture, loss function, and regularization should be used; iii) with the choice in i) and ii), whether the trained estimator approximates MLE and performs in a robust and workload-agnostic way.

\sstitle{MLE formulation.} We first formulate NDV estimation as an MLE problem. Assuming that a data column is generated from some prior distribution and a sample $\sample$ is drawn uniformly at random from the column, we observe the sample profile, \ie, a vector $f = (f_j)_{j=1, \ldots}$ where $f_j$ is the number of distinct values with frequency $j$ in $\sample$. For example, for $\sample = \{{\rm a, a, a, b, b, b, c, c}\}$, we have $f_1 = 0$, $f_2 = 1$ (`$\rm c$' has frequency $2$), and $f_3 = 2$ (`$\rm a$' and `$\rm b$'). For an MLE estimation, we aim to find the value of NDV, $D$, such that the probability of observing $f$ conditioned on ${\rm NDV} = D$ is maximized. Although it is difficult to solve (even approximately) the MLE formulation, it will guide the design and analysis of our machine learning model.

\sstitle{Learning estimator that approximates MLE.} With the sample profile $f$ as a feature and the ratio error between the estimate and the true NDV as the training loss, we have a skeleton of the learning framework.
%
%
Unlike learning MLE for mean and STD where the training data preparation is trivial because each dimension of the feature space ($v_i$ in $\sample$) can be independently generated, in NDV estimation, however, different dimensions of the sample profile ($f_i$ in $f$) are correlated, conditioned on NDV; thus, it is more difficult to characterize the distribution needed for preparing the training set. 
After carefully investigating the probability distribution the model learns based on our choice of feature and loss and comparing the learned probability distribution with the one in our MLE formulation, we derive several precise properties to characterize the training distribution we need in order to make the learned estimator approximate an MLE estimator.

\sstitle{Efficient training data generation.} We design an efficient algorithm to generate training data points. To generate one data point, we need to first generate a data column (in a compact way). And instead of drawing a sample explicitly, we directly generate the sample profile from its distribution determined by the data column. In our experiment, $10^5$-$10^6$ data points suffice to train the estimator.

\sstitle{Instance-wise negative results and model regularization.} It is important to note that the negative result about sample-based NDV estimation in \cite{charikar2000towards} also hold for ``learned estimators''. However, the negative result in \cite{charikar2000towards} is a global one, in the sense that there exist {\em hard instances} of data columns whose NDVs are difficult to be estimated. However, the hope is that, upon the observation of the sample profile, it is possible to infer that the underlying data column is not a hard instance and an estimate with error lower than the global lower bound can be expected. We first derive a new instance-wise negative result, \ie, a lower bound of the estimation error upon the observation of a sample profile. We use this negative result to regularize the training of our model. More specifically, we propose a regularization method that encourages the learned estimator to achieve just the instance-wise lower bound of estimation error, instead of always minimizing the estimation error to be zero (which is an impossible goal due to global negative result). Its effectiveness will be verified via experimental studies.

\sstitle{Deployment and experiments.} The learned estimator, once trained, can be deployed in any data warehouse without the need of re-training or tuning. We publish our estimator online \cite{url:est} as a trained neural network that takes a sample/sample profile of a column as the input feature and outputs an estimated NDV of that column. It is compared with the state-of-the-art sample-based NDV estimators and is shown to outperform them on nine different real datasets. Note, again, that none of these datasets is used to train our estimator, and all the training data is generated synthetically.

\subsection*{Related Work}
\noindent\textbf{Learned cardinality estimation.} There have been a long line of existing works on learning a model to estimate cardinality or selectivity of a query~\cite{anagnostopoulos2015learning, lakshmi1998selectivity, liu2015cardinality, dutt2019selectivity, kipf2018learned,pvldb:DuttWNC20,vldb:ZhuW21}. In general, they can be divided into two types~\cite{wang2020we}.
The first type uses query as the primary feature and training data comes from the records of query executions.
For example, \cite{dutt2019selectivity} uses query as features and applies tree-based ensembles to learn the selectivity of multi-dimensional range predicates. This type of method typically requires executing a huge number of queries to obtain enough training data~\cite{wang2020we}. There are efforts to alleviate this issue: \cite{pvldb:DuttWNC20} proposes to reduce model construction cost by incrementally generating training data and using approximate selectivity label. 
Generally, this type of methods work well when future queries follow the same distribution and similar templates (\eg, conjunction of range and categorical predicates) as the training data~\cite{vldb:ZhuW21}.
The second type (for example \cite{hilprecht2020deepdb} and \cite{vldb:ZhuW21}) builds a model to approximate the joint distribution of all attributes in a table. This type of methods 
require re-training in case of data or schema update.
In dynamic environments with unseen datasets and workloads coming, both type of methods 
require huge efforts of accessing the new dataset and retraining~\cite{wang2020we}.

Although we also learn a model to estimate NDV/cardinality, our method is completely different from the above methods. Our method is a sample-based approach and we use features from sample rows (drawn from tables or query results) instead of queries. More importantly, our method is workload agnostic (only needing to be trained once) and can be applied to any dynamic workload.


\noindent\textbf{Sample-based NDV estimation.}
Existing sampling based methods are typically constructed based on heuristics or derived by making certain assumptions.
For example, there are estimators derived by assuming infinite population size~\cite{chao1992estimating}, certain condition of skewness~\cite{shlosser1981estimation}, and certain distribution of the data~\cite{motwani2006distinct}.
GEE estimator \cite{charikar2000towards} is constructed to match a (worst-case) lower bound of estimation error in \cite{charikar2000towards}.
%
%
We discuss these existing approaches in more details in \csec\ref{ssec:existing_est}. These approaches are not robust especially on datasets where their assumptions about data distribution break, as we will show in experiments in \csec\ref{sec:exp}. 

\noindent\textbf{Sketch-based NDV estimation.} Sketch-based methods (\eg, Hyperloglog~\cite{flajolet2007hyperloglog}) scan the entire dataset once and keep a memory-efficient sketch that is used to estimate NDV. 
\revise{These methods are able to produce highly accurate NDV estimation. For example, the expected relative error of HyperLogLog ~\cite{flajolet2007hyperloglog} is about $1.04/\sqrt{m_{\text{bytes}}}$, 
%
%
%
where $m_{\text{bytes}}$ is the number of bytes used in HyperLogLog; With 10K bytes, the relative error of NDV estimation can be as small as $1\%$-$2\%$.
See survey~\cite{harmouch2017cardinality} for a comprehensive review.
}

\noindent\revise{\noindent\textbf{Sample v.s. Sketch.} When one scan of the data is affordable, sketch-based estimators are preferable to sample-based ones.}
%
\revise{
In scenarios where a full scan of the data is (relatively) too expensive, sample-based methods are preferable. For example, when NDV estimation in a column is used by the query optimizer to generate a good execution plan for a SQL query, a high-quality sample-based estimation is preferable, as the query is almost processed after a full scan. Another example is approximate query processing \cite{sigmod:ChaudhuriDK17}, where offline samples are used to provide approximate answers to analytical (\eg, NDV) queries over voluminous data with interactive speed. In short, sketch-based estimators and sample-based methods are two orthogonal lines of research with different applications.
%
}


\subsection*{Paper Organization}
%
In \csec\ref{sec:preliminary}, we provide the preliminaries including an MLE-based formulation for NDV estimation, and review some representative estimators.
In \csec\ref{sec:overview}, we provide an overview of our learning-to-estimate framework; we analyze and derive properties of the training distribution needed to make the learned model approximate MLE.
In \csec\ref{sec:learning}, we give details about training data generation and model architecture; we also introduce our new instance-wise negative result about sample-based NDV estimation, and how to regularize the model accordingly. 
\csec\ref{sec:use} gives a brief introduction on how to use our learned estimator that is available online.
We evaluate our estimator and compare it with the state of the arts experimentally and report the results in \csec\ref{sec:exp}.

%% file: paper-pre.tex

\section{Preliminaries}
\label{sec:preliminary}



We focus on a specific {\em data column} $\col$ of a table with $N$ rows. $N$ is called the {\em population size}. Let $D$ be the {\em number of distinct values} (NDV) in $\col$. When calculating NDV of the column $\col$, we consider $\col$ as a (multi)set of values from some (possibly infinite) domain $\dom$. 
%

%
\stitle{Problem statement: estimating NDV from samples.}
We want to estimate $D$ from a random sample $\sample \subseteq \col$ of $n$ tuples drawn uniformly at random from $\col$. Let $r = n/N$ be the sampling rate. We assume $N$, or equivalently $r$, is observed.

We first formally define two important notations.

%


%
{\bf Frequency.} The {\em frequency} of a value $i \in \dom$ in a column $\col$, or a sample $\sample$, is the number of times it appears in $\col$, or $\sample$, denoted as $N_i$, or $n_i$, respectively. By definitions, we have $\sum_i N_i = N$ and $\sum_i n_i = n$, and the NDV in $\col$ is $D = |\{i\in\Omega \mid N_i>0\}|$. 

{\bf Profile.} In order to calculate NDV, it is {sufficient} to consider the {\em profile} of $\col$, denoted as $F = (F_j)_{j=1, \ldots, N}$, where $F_j = |\{i \in \dom \mid N_i = j\}|$ is the number of distinct values with frequency $j$ in $\col$; similarly, the profile of a random sample $\sample$ is $f = (f_j)_{j = 1, \ldots, n}$ where $f_j = |\{i \in \dom \mid n_i = j\}|$ is the number of distinct values with frequency $j$ in $\sample$.
By definitions, the NDV in $\col$ is $D = \sum_{j>0}F_j$ and the population size is $N=\sum_j jF_j$; The NDV in sample $\sample$ is $d = \sum_{j>0}f_j$ and the sample size is $n=\sum_j jf_j$.


\subsection{An MLE-based Formulation.}
Estimating NDV from random samples can be formulated as a {\em maximum likelihood estimation} (MLE) problem, which is commonly used to estimate unknown parameters of a statistical model, with desirable properties such as consistency and efficiency \cite{chambers2012maximum}. 
The estimated value (or the solution to an MLE problem) maximizes the probability of the observed data generated from this model.

We assume that the column $\col$ with profile $F$ is drawn from some prior probability distribution. 
A uniformly random sample $\sample$ with profile $f$ is then drawn from $\col$.
An MLE-based formulation for NDV estimation can be derived based on the {\em observed} profile in the sample $\sample$, \ie, the sample profile $f$, and
%
%
%
the observed population size $N$ (or equivalently, sampling rate $r=n/N$).
We estimate $D$ as the one that maximizes the probability of observing $f$ and $N$.
\[\label{equ:dmle:general}
D^{\sf MLE} = \argmax_{D} \pr{f,N \mid D} = \argmax_{D} \sum_{F} \pr{f,N \mid F}\pr{F \mid D}.
\]
%
%
%

Define $\trainprof(D, N) = \{F \mid \sum_{j>0} F_j = D ~\text{and}~ \sum_{j>0} j \cdot F_j = N\}$ to be all the {\em feasible} profile configurations with NDV equal to $D$ and population size equal to $N$. For $F \in \trainprof(D, N)$, we have $\pr{f,N \mid F} = \pr{f \mid F}$; and for $F \notin \trainprof(D, N)$, we have $\pr{f,N \mid F} = 0$ and $\pr{F \mid D} = 0$. The above formulation can be rewritten as:
\[\label{equ:dmle:general2}
D^{\sf MLE} = \argmax_{D} \!\!\! \sum_{F \in \trainprof(D,N)} \!\!\! \pr{f \mid F} \pr{F \mid D}.
\]

The estimator $D^{\sf MLE}$ is to be used on unknown columns; so in order to solve the above optimization problem, it is reasonable to assume that the prior distribution of $F$ is uniform, in the sense that every possible profile in $\trainprof(D, N)$ appears with equal probability, \ie, $\pr{F \mid D} = 1/|\trainprof(D, N)|$ for every $F \in \trainprof(D, N)$. Under this assumption, we want to solve the following one for $D^{\sf MLE}$:
\begin{equation}\label{equ:dmle}
D^{\sf MLE} = \argmax_{D} \frac{1}{|\trainprof(D, N)|}\sum_{F \in \trainprof(D,N)} \!\!\! \pr{f \mid F}.
\end{equation}
We can also interpret the MLE-based formulation \eqref{equ:dmle} as follows. After observing the sample profile $f$ and the population size $N$, we estimate $D$ as $D^{\sf MLE}$ which maximizes the average probability of generating $f$ from a feasible profile $F \in \trainprof(D, N)$.
%
Solving \eqref{equ:dmle}, however, is difficult even approximately, and thus this formulation has not been applied for estimating NDV yet.


%

%

A natural question is whether it benefits to use sample $\sample$, instead of sample profile $f$, as observed data to derive MLE and as features in our learning framework.
\if\TR 1
We provide a formal analysis on why using sample is not more advantageous in Appendix~\ref{ssec:sample_vs_profile}.
\else
We defer a formal analysis on why using sample is not more advantageous to a technical report~\cite{url:technical_report} due to space limit.
\fi
In fact, most existing estimators (as we show next) also use sample profile instead of sample to estimate NDV. 
\subsection{Existing Estimators}
\label{ssec:existing_est}
There have been a long line of works on estimating NDV from random samples. We review some representative ones as follows.
\begin{itemize}[wide, labelindent=0pt]
\item A problem related to the one in \eqref{equ:dmle} is {\em profile maximum likelihood estimation} (PML) \cite{hao2019broad,pavlichin2019approximate,charikar2019efficient}, which chooses $F$ that maximizes the probability of observing $f$ of the randomly drawn $\sample$. Define:
\begin{equation}\label{equ:fpml}
F^{\sf PML} = \argmax_F \pr{f \mid F} ~~~~~\text{~~~~~~and~~~~~~}~~~~~ \hbox{$D^{\sf PML} = \sum_{j>0} F^{\sf PML}_j$}.
\end{equation}
There have been works on finding approximations to $F^{\sf PML}$ \cite{pavlichin2019approximate, charikar2019efficient}, which can be in turn used to obtain an approximate version of $D^{\sf PML}$ in \eqref{equ:fpml}, although, in general, $D^{\sf MLE} \neq D^{\sf PML}$. 
\item {\sf Shlosser}~\cite{shlosser1981estimation} is derived based on an assumption about skewness: $\epinline{f_i}/\epinline{f_1} \approx F_i/F_1$, and performs well when each distinct value appear approximately one time on average~\cite{haas1995sampling}.  It estimates $D$ as
\begin{equation} \label{equ:shlosser}
\hbox{$D^{\sf Shlosser} = d + (f_1 \sum_{i=1}^n (1-r)^i f_i) \Big/ (\sum_{i=1}^n i r (1-r)^{i-1} f_i)$}.
\end{equation}
\item {\sf Chao}~\cite{chao1984nonparametric} approximates the expected NDV, $\epinline{D}$, in large population for some underlying distribution, and estimates NDV as a lower bound of $\epinline{D}$ with the population size approaching infinity:
\begin{equation} \label{equ:chao}
D^{\sf Chao}=d+f_1^2/(2f_2).
\end{equation}
%
%
%
\item {\sf GEE}~\cite{charikar2000towards} is constructed by using geometric mean to balance the two extreme cases for values appearing exactly once in the sample: those with frequency one in $\col$ and drawn into $\sample$ with probability $r$ v.s. those with high frequency in $\col$ and at least one copy drawn into $\sample$. It is proved to match a theoretical lower bound of ratio error for NDV estimation within a constant factor.
\begin{equation} \label{equ:gee}
\hbox{$D^{\sf GEE} = \sqrt{1/r} \cdot f_1+\sum_{i=2}^nf_i$}
\end{equation}
{\sf HYBGEE}~\cite{charikar2000towards} is a hybrid estimator using {\sf GEE} for high-skew data and using the smoothed jackknife estimator for low-skew data. 
{\sf AE}~\cite{charikar2000towards} is a more principled version of {\sf HYBGEE} with smooth transition from low-skew data to high-skew data. It requires to solve a non-linear equation using, \eg, Brent’s method~\cite{brentq}.
\end{itemize}
\subsection{Negative Results}\label{sec:pre:ndv:negative}
The aforementioned lower bound of ratio error for sample-based NDV estimators is given by Charikar \etal in \cite{charikar2000towards}. 
More formally, define the {\em ratio error} of an estimation ${\hat D}$ w.r.t. the true NDV $D$ to be
\begin{equation} \label{equ:errdef}
\err({{\hat D}},D) = \max\{{{\hat D}}/D, D/{{\hat D}}\}.
\end{equation}
It considers in \cite{charikar2000towards} a even larger class of estimators which randomly and adaptively examine $n$ tuples from $\col$. Note that drawing a random sample $\sample \subseteq \col$ of $n$ tuples is a special case here.
It says that for any such estimator, there exists a choice of column $\col$ such that, with probability at least $\gamma > e^{-n}$, the ratio error 
is at least
\begin{equation} \label{equ:hardness}
\hbox{$\err({{\hat D}},D) \geq \sqrt{\frac{N-n}{2n}\ln\frac{1}{\gamma}}$}.
\end{equation}

%

%% file: paper-overview.tex

\section{Overview of Learning Framework}
\label{sec:overview}
Since analytically solving \eqref{equ:dmle} even approximately is difficult, we propose to formulate the task of deriving $D^{\text{MLE}}$ as a supervised learning problem.
We first introduce some key elements in the learning model, including the loss function and the design of training dataset, which learns an estimator to approximate \eqref{equ:dmle}.

\subsubsection*{Learning to estimate}
%
%
In typical traditional estimators such as \eqref{equ:shlosser}-\eqref{equ:gee}, the NDV is estimated as a {\em function} ${\hat D} = \est(f, r)$ of the sample profile $f$ and the sampling rate $r$, or equivalently a {\em function} ${\hat D} = \est(f, N)$ of $f$ and the population size $N$ as $r =  \sum_{i>0}if_i/N$. 
%
%
As it is difficult to directly derive the function $h$ from a principled formulation such as MLE in \eqref{equ:dmle}, we attempt to {\em learn} the MLE of NDV from a set of training data points $A = \{((f,N), D)\}$, where in each point, $(f,N)$ is the {\em input feature} and the NDV $D$ is the {\em label} to be predicted. We use the {ratio error} defined in \eqref{equ:errdef} to measure the accuracy of estimations, and accordingly, the loss function of the model $\est$ is:
\begin{equation} \label{equ:loss}
L(\est(f, N)={\hat D}, D) = |\log {\hat D} - \log D|^2 = \left(\log\err({\hat D},D)\right)^2.
\end{equation}
%
%
Given a hypothesis set $\estclass$ of estimation functions, the goal is to find $\est \in \estclass$ with small {\em empirical loss}:
\begin{equation} \label{equ:totalloss}
\hbox{$\hat R_A(\est) = 1/|A| \sum_{((f,N), D) \in A} L(\est(f, N), D)$}.
\end{equation}

In order to generate a data point $((f,N), D)$ in the training set $A$, we first generate a data column $\col$ with profile $F$ according to a {\em training distribution} $\traindist$. The NDV $D$ as well as population size $N$ is directly calculated from $F$. A random sample $\sample$ is drawn uniformly at random from $\col$ with sampling rate $r$. The sample profile $f$ is then obtained from $\sample$. The above process is repeated independently multiple times to generate a training set $A = \{((f,N), D)\}$.
\revision{We only generate one sample profile $f$ from a profile $F$ to ensure that each training data point in $A$ is independently and identically distributed.}


\subsubsection*{Approximating MLE}
\label{sssec:workload_agnostic}
Intuitively, for a different training distribution $\traindist$, a different learned estimator $\est$ will be derived from the hypothesis set (model architecture) $\estclass$. Assuming that $\estclass$ is expressive enough \cite{icml:RaghuPKGS17}, we give some guidelines here on how to choose $\traindist$ so that the learned $\est$ approximates the MLE estimator $D^{\sf MLE}$.


Let $\prp{N, D, F, f}$ denote the joint distribution of $(N, D, F, f)$ in the training set $A$ that is generated from the training distribution $\traindist$ as above. 
%
%
Using the loss function in \eqref{equ:loss}-\eqref{equ:totalloss}, \ie, the L2 loss on $\log D$, the trained model $\est$ learns the distribution $\prp{\log D \mid f,N}$~\cite{goodfellow_bengio_courville_2017}. By minimizing \eqref{equ:totalloss}, the model tends to emit an output
\begin{equation}
\label{eq:model_approx_data_distribution}
h(f,N) \approx \arg \text{max}_{D} \prp{\log D \mid f,N},
\end{equation}
for a given input $(f,N)$. The approximation sign $\approx$ is because the trained model might not be able to learn the underlying distribution $\traindist$ of the training set exactly and the accuracy of the approximation is also determined by the hypothesis set $\estclass$ and the training algorithm.
For the term $\prp{\log D \mid f,N}$ on the right hand side,
\begin{align}
& \prp{\log D \mid f,N} = 1/\prp{f \mid N}\cdot \prp{\log D \mid N} \prp{f \mid \log D,N} \nonumber \\
& = \frac{\prp{\log D \mid \log N}}{\prp{f \mid N}}\sum_{F \in \trainprof(D,N)} \!\!\!\!\pr{f \mid F}\prp{F \mid D,N} \label{equ:trainobj}
\end{align}
where both equations are from properties of conditional probability.

Consider the following two conditions about the distribution $\traindist$:
\begin{align}
& \hbox{i) for any $N$,~~} \prp{\log D \mid \log N} = \text{constant}; \label{eq:training_data_requirement1}
\\
& \hbox{ii) for any $N$ and $D$,~~} \prp{F \mid D,N}=1/|\trainprof(D,N)|. \label{eq:training_data_requirement2}
\end{align}
Namely, i) for any fixed $N$, the NDV $D$ distributes uniformly at log scale (or equivalently $\log D$ distributes uniformly) in $\traindist$; and ii) for any fixed $N$ and $D$, every feasible profile appears with equal probability. If both are satisfied, the maximizer of \eqref{equ:trainobj} (as a function of $D$) can be written as
%
\begin{small}
\begin{align}
 &\argmax_{D} \prp{\log D \mid f,N} \nonumber
= \argmax_{D} \frac{\hbox{constant}}{\prp{f \mid N}}\sum_{F \in \trainprof(D,N)} \frac{\pr{f \mid F}}{|\trainprof(D,N)|}\nonumber
\\
& =  \argmax_{D}\frac{1}{|\trainprof(D,N)|}\sum_{F \in \trainprof(D,N)} \pr{f\mid F} = D^{\sf MLE} \label{equ:trainobjmax}
\end{align}
\end{small}
by putting \eqref{eq:training_data_requirement1}-\eqref{eq:training_data_requirement2} back into \eqref{equ:trainobj}. The second equality in \eqref{equ:trainobjmax} is because $\prp{f \mid N}$ is a probability term that is independent on $D$; and the last equality is from the definition of $D^{\sf MLE}$ in \eqref{equ:dmle}. Therefore, the learned estimator $\est(f,N)$ approximates $D^{\sf MLE}$ if the training distribution satisfies the two conditions in \eqref{eq:training_data_requirement1}-\eqref{eq:training_data_requirement2}.


From the above analysis, the learned estimator $\est(f,N)$ approximates $\eqref{equ:trainobjmax}$ which depends on only $\trainprof(D,N)$ and $\pr{f \mid F}$: the former is a deterministic finite set, and the latter is a probability distribution depending only on the sampling procedure but not on $\traindist$; therefore, intuitively, we can expect that $\est(f,N)$ is workload-agnostic, \ie, it generalizes well on unseen data columns, which will be verified later in our experiments reported in \csec\ref{sec:exp}. 

\noindent\revise{\noindent{\bf About uniformity of $F$.}
When introducing the MLE estimator $D^{\sf MLE}$ and analyzing its equivalence to the learned model $\est(f,N)$, the underlying assumption is that the prior distribution of $F$ is uniform. In fact, any targeted prior distribution of $F$ can be plugged into $D^{\sf MLE}$ and accordingly, used as the training distribution $\traindist$ for $\est$. However, we assume the uniformity purposely to enable the learned estimator generalize to unseen datasets (workload-agnostic). We tried to train an estimator ({\em Lower Bound (LB)} in \csec\ref{sec:exp}) with the prior of $F$ and $\traindist$ the same as those in the test datasets. 
In experiments, we observed that the LB estimator works well only on this particular workload but does not generalize to unseen workloads. In comparison to the LB estimator with the ``optimal'' prior, our workload-agnostic estimator with the uniform prior has comparable performance on every test table (refer to, \eg, \ctab\ref{tbl:overall_performance}).}

%% file: paper-method.tex
\section{Learning Estimator from Data}
\label{sec:learning}


\subsection{Efficient Training Data Generation}
\label{sec:learning:trainingdata}
As long as the training data satisfies the conditions in \eqref{eq:training_data_requirement1}-\eqref{eq:training_data_requirement2}, whether the data is synthetic or from real world makes no difference. This makes it possible for us to train the model using synthetically generated data without needing any real-world data.

To generate a training data point $((f,N), D)$, one may first generate a random column $\col$, then draw a random sample $\sample$ from $\col$ and calculate sample profile $f$ from $\sample$. However, since the population profile $F$ of $\col$ contains all required information to generate a random sample profile $f$ (as we will show in Section~\ref{sssec:sample_profile_gen}), we can directly randomly generate a population profile $F$ and then draw a random sample profile $f$ from $\prp{f|F}$.


\subsubsection{Population profile generation}
\label{sssec:pop_profile_gen}
Generating population profile is to draw samples from the distribution $\prp{F}$, written as:
\begin{equation}
\hbox{$\prp{F}= \sum_N\sum_D \prp{F|N,D}\prp{D|N}\prp{N}$}.
\end{equation}
We can draw samples from $\prp{F}$ by repeating the following: sample a $N$ from $\prp{N}$, sample a $D$ from $\prp{D|N}$, and sample a $F$ from $\prp{F|N,D}$, with
$\prp{D|N}$ and $\prp{F|N,D}$ satisfying the conditions in \eqref{eq:training_data_requirement1}-\eqref{eq:training_data_requirement2}.
%
%
In order to ensure the training data to be diverse, we want to have $N$ cover a big range of magnitude, so we set the distribution of $N$ to be uniform at log scale, \ie, $\log_{10} N \sim U(0,B)$ where $B$ is a constant specifying the maximum population size. We select $B=9$ according to the memory limit of our machine. 

Generating $F$ from $\prp{F|N,D}$ is, however, non-trivial.
Given population size $N$ and population NDV $D$, drawing a population profile from $\prp{F|N,D}$ is equivalent to randomly sampling an element under i) the uniform-distribution constraint (\ie, every feasible $F$ has equal probability to be drawn) from set $\trainprof(D, N)$ which is specified by the two constraints ii) $\sum_{j>0} F_j = D$ and iii) $ \sum_{j>0} jF_j = N$. Designing such a sampling procedure that satisfies the three constraints simultaneously is very challenging.
%

\begin{figure}
\Description{}
  \centering
  \includegraphics[width=0.9\linewidth]{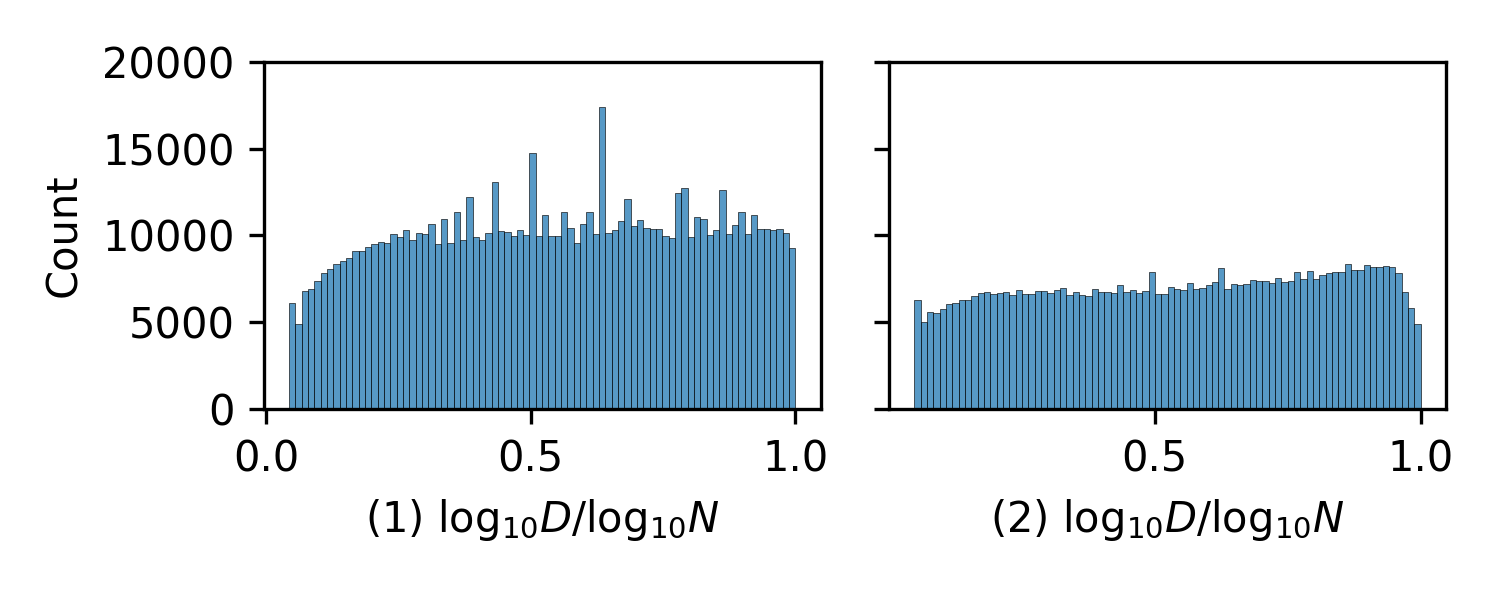}
  \vspace{-2mm}
  \caption{Frequency distribution of ${\log D}/{\log N}$.}
  \label{fig:log_D_div_N_distri}
\end{figure}

If the dimensionality of $F$ 
can be given, the sampling procedure is easier.
Consider an alternative form of $\prp{F}$:
\begin{equation}
\label{eq:gen_F_by_N_M}
\hbox{$\prp{F} = \sum_N \sum_M \prp{F|N,M}\prp{M|N}\prp{N}$}.
\end{equation}
This indicates an alternative procedure: sample a $N$ from $\prp{N}$, sample a $M$ from $\prp{M|N}$, and sample a $F$ from $\prp{F|N,M}$.
In this procedure, dimensionality $M$ is given when sampling $F$ from distribution $\prp{F|N,M}$. The distributions $\prp{D|N}$ and $\prp{F|N,D}$ are now implicitly induced \revise{from $\prp{F}$}.
\revise{To see why it is the case, when $\prp{F}$ is derived as in \eqref{eq:gen_F_by_N_M}, the joint distribution $\prp{F,N,D}$ is determined, because, by definitions, $N$ and $D$ are deterministically dependent on $F$, \ie, $\prp{F,N,D}=\prp{F}$ if $N=\sum_i iF_i$ and  $D=\sum_i F_i$, and $\prp{F,N,D}=0$ otherwise. $\prp{N,D}$ and $\prp{N}$ are just marginal distributions of $\prp{F,N,D}$. Therefore, by the definition of conditional probabilities, we can obtain $\prp{D|N}$ as $\prp{N,D}/\prp{N}$ and $\prp{F|N,D}$ as $\prp{F,N,D}/\prp{N,D}$.}

We plan to design the two distributions $\prp{M|N}$ and $\prp{F|N,M}$ to make the induced $\prp{D|N}$ and $\prp{F|N,D}$ approximately satisfy the conditions in \eqref{eq:training_data_requirement1}-\eqref{eq:training_data_requirement2}. 
In the following, we first intuitively specify the form of $\prp{M|N}$ and $\prp{F|N,M}$, then derive an efficient sampling algorithm to generate $F$, and show that the conditions in \eqref{eq:training_data_requirement1}-\eqref{eq:training_data_requirement2} are approximately satisfied in the end.

\noindent{\bf Achieving (approxiamte) uniformity in sample.}
Intuitively, $M$ is the highest number of times that a value can appear in population.
We draw $M$ uniformly at log scale \ie $\log_{10} M \sim U(0,B)$. Note that $M$ is the dimensionality of the space where we draw $F$ and it is different from the actual length of $F$ (maximum $l$ with $F_l>0$), so $F_M$ can be zero.
Thus, $M$ and $N$ are independent (with $\log_{10} N \sim U(0,B)$), and $\prp{M|N}$ is identical to $\prp{M}$.

Given $N$ and $M$, similarly we can have a feasible configuration set for $F$: $\trainprofp(N,M) = \{F|\sum_{i=1}^M iF_i=N\}$.
To make $\prp{F|D,N}$ approximately uniform (required in \eqref{eq:training_data_requirement2}), intuitively, we need to make the distribution of $F$ as diffusive as possible, so we also draw $F$ uniformly from $\trainprofp(N,M)$.
Notice that compared with $\trainprof(N,D)$, $\trainprofp(N,M)$ only has one constraint and the dimensionality of $F$ is given. This make it much easier to design a sampling procedure. 

Let $SF_i$ denote $\sum_{i}^M F_i$, then $F_i = SF_i-SF_{i+1}$. Apparently, $SF$ has a one-to-one mapping relationship with $F$, so drawing each feasible $F$ with equal probability is equivalent to drawing each feasible $SF$ with equal probability. The feasible set for $SF$ is $    Q(N,M) = \{SF \mid \sum_{i=1}^M SF_i = N;\  SF_i\geq SF_{i+1}\  \forall\  i\}$.
Without the constraint $SF_i\geq SF_{i+1}$, the problem of generating $SF$ so that each feasible $SF$ has equal probability to be generated is known as the \textit{random fixed sum} problem, and there are existing efficient algorithms to solve it in $O(M\log M)$~\cite{random_fixed_sum, rfs_matlab}. The constraint $SF_i\geq SF_{i+1}$ can be easily satisfied afterwards by reassigning $SF$ to be its sorted version. Once a $SF$ is generated, the corresponding $F$ can be obtained.

To summarize, the process of generating one population profile $F$ is as follows: draw $N$ from $\log_{10} N \sim U(0,B)$, draw $M$ from $\log_{10} M \sim U(0,B)$, draw $SF$ from $Q(N,M)$ using algorithms for the random fixed sum problem, and obtain $F$ by $F_i = SF_i-SF_{i+1}$. 


As discussed above, $\prp{F|N,D}$ and $\prp{\log D|\log N}$ can be induced from the distribution $\prp{F}$, and thus also from $\prp{F|N,M}$ and $\prp{M|N}$. The uniformity of $\prp{F|N,D}$ and $\prp{\log D|\log N}$ is still not strictly guaranteed. For $\prp{F|N,D}$ and $\prp{F|N,M}$, however, the sizes of the supporting sets $\trainprof$ and $\trainprof'$, respectively, are both extremely big and it can be shown the fraction of $F$ that we can sample with our best effort ($\sim10^9$ elements of $F$) is at the scale of $10^{-3991}$.
This leads to the fact that the samples we drawn are extremely sparse in the space of all possible $F$ and each $F$ can at most appear once in the samples. Therefore, the empirical distribution of $\prp{F|N,D}$ in training data will be close to uniform. 
For $\prp{\log D|\log N}$, we empirically show its uniformity in \cfig\ref{fig:log_D_div_N_distri}. For $N$ and $D$ drawn from $\log_{10} N\sim U(0,B)$ and $\log_{10} D \sim U(0,\log_{10} N)$, the ratio
$\log D/\log N$ should follow the uniform distribution $U(0,1)$ and we show the frequency distribution of $\log D/\log N$ in \cfig\ref{fig:log_D_div_N_distri}(1); 
we obtain the $N$ and $D$ of our generated profiles and plot the frequency distribution of $\log D/\log N$ -- as shown in \cfig\ref{fig:log_D_div_N_distri}(2), it is also roughly uniformly distributed.

\subsubsection{Sample profile generation}
\label{sssec:sample_profile_gen}
To draw a sample, we need to know the sampling rate $r$. The conditions in \eqref{eq:training_data_requirement1}-\eqref{eq:training_data_requirement2}
have no constraint on the distribution of $r$. To ensure the training data to be diverse, we would like to have $r$ cover a big range of magnitude. Therefore, we draw sampling rate $r$ uniformly at log scale by $\log_{10} r \sim U(-B', -1)$ where we select $B'=4$ as $10^{-4}$ is a small enough sampling rate in practice.

Instantiating the population/column from $F$ and then performing random sampling to get a sample $\sample$ and then sample profile $f$ is of complexity $O(N)$. We propose to directly perform sampling from $F$ with a complexity of $O(D)$.
When performing sampling, for a value appeared $K$ times in population, the number of times $k$ it appears in sample follows a binomial distribution $\pr{k} = \binom{K}{k} r^k(1-r)^{K-k}$
where $r$ is the sampling rate. By performing a binomial toss for every distinct value in population, we obtain the number of times each value appear in sample and sample profile $f$ can be calculated. Accordingly, the complexity of generating a sample profile is $O(D)$.


%

\subsubsection{Diversity training data}
\label{sssec:diversify}
To further improve the generalization ability of the trained model, we diversify the profiles we draw from $\trainprofp$ by incorporating some human knowledge. This is in the same spirit as that one may incorporate human knowledge to diversify an image dataset by operations such as rotation to improve the generalization ability of models trained on the dataset~\cite{shorten2019survey}.

We diversify our training data based on the following intuition:
each $F$ can be seen as a point in high dimensional space with $F_i$ being its coordinate at the $i$th dimension; we want to enlarge the supporting region of the points (\ie, the region that the points span) so that the trained model generalizes better,
since machine learning models are better at interpolation than extrapolation~\cite{xu2020neural}.
%

The most ideal way to increase the support region of the profiles is to have some data points with a much bigger population size. In this way, $F_i$ at every dimension can be very big yielding a bigger support region. However, due to hardware limit in practice we are not able to achieve this, so we choose to randomly increase one $F_i$ along one single random dimension for each $F$. 
Specifically, for each profile $F$ generated by the method in Section~\ref{sssec:pop_profile_gen}, 
a new component $F'$ is added to $F$ to obtain 
the final profile $F''=F+F'$ where $F'$ contains only one positive value $F'_{i_{\text{p}}}=D'$ and $F'_{i\neq i_{\text{p}}} = 0$;
$D'$ and $i_{\text{p}}$ are randomly generated by $\log_{10}N' \sim U(0,B) \ ,\ \log_{10}D' \sim U(0,\log_{10}N')$, and $\ i_{\text{p}}=\nint{N'/D'}$. Since $F'$ only have one non-zero value $F'_{i_{\text{p}}}$, with the same population size distribution, $F'_{i_{\text{p}}}$ will be much greater than $F_{i_{\text{p}}}$ on average. In this way, the supporting region of the profiles $F$ used in training is increased greatly.

\subsection{Feature Engineering and Model Structure}
\label{sec:learning:model}

\subsubsection{Feature engineering}
%
%
%
%
%



The raw features include sample profile $f$ and population size $N$. 
On top of the two raw features, other meaningful features that we can derive include sample size $n=\sum_i if_i$, sample NDV $d=\sum_i f_i$, and sampling rate $r = n/N$.

The number of elements in the sample profile $f$ varies for different samples. The length of $f$ can be as large as the sample size and as be as small as one. However, during model training, we have to use a fixed number of features. We choose to keep only the first $m$ elements of the sample profile $f$, \ie $f[1:m]$. \revise{If the length of $f$ is less than $m$, we pad it with zeros.} This is based on the intuition that the predictive power of $f_i$ decreases as $i$ increases. In fact, some of the well-known estimators only use the first few elements in the sample profile $f$, yet achieving fairly good performance~\cite{charikar2000towards,chao1984nonparametric,haas1995sampling}.
To make up for the elements being cutoff in the sample profile $f$, we add the corresponding {\em cut-off sample size} $n_c=\sum_{i=m+1} if_i$ and {\em cut-off sample NDV} $d_c = \sum_{i=m+1} f_i$ as two additional features.

The feature set we use is $x = \{N, n, n_c, d, d_c, 1/r, f_1, \dots, f_m \}$, with a total of $N_x = m+6$ features, and the single target we aim to predict is population NDV $D$. We set $m = 100$ by default in experiments.

\subsubsection{Model structure}
\label{sssec:model_struc}
\revise{
Recall that our goal is to learn an estimator/model to approximate $D^{\sf MLE}$. Simple models often assume some specific relationship between the input features, \ie, the sample profile $f$ in our case, and the label to be predicted, \ie, NDV $D$. For example, linear/logistic regression assumes a linear relation between $f$ and (transformed) $D$: \eg, $D = \sum_i a_i f_i$. However, $D^{\sf MLE}$ can be any {\em unknown} function of $f$ (\eg, refer to the previous estimator $D^{\sf Shlosser}$ introduced in \csec\ref{ssec:existing_est}) that is much more complex than the above restricted class of linear functions.
Therefore, since neural networks are able to approximate any function (with enough nodes and layers)~\cite{haykin1998nn}, we choose our model to be a neural network.}

\begin{figure}[t]
\Description{}
  \centering
  \includegraphics[width=0.8\linewidth]{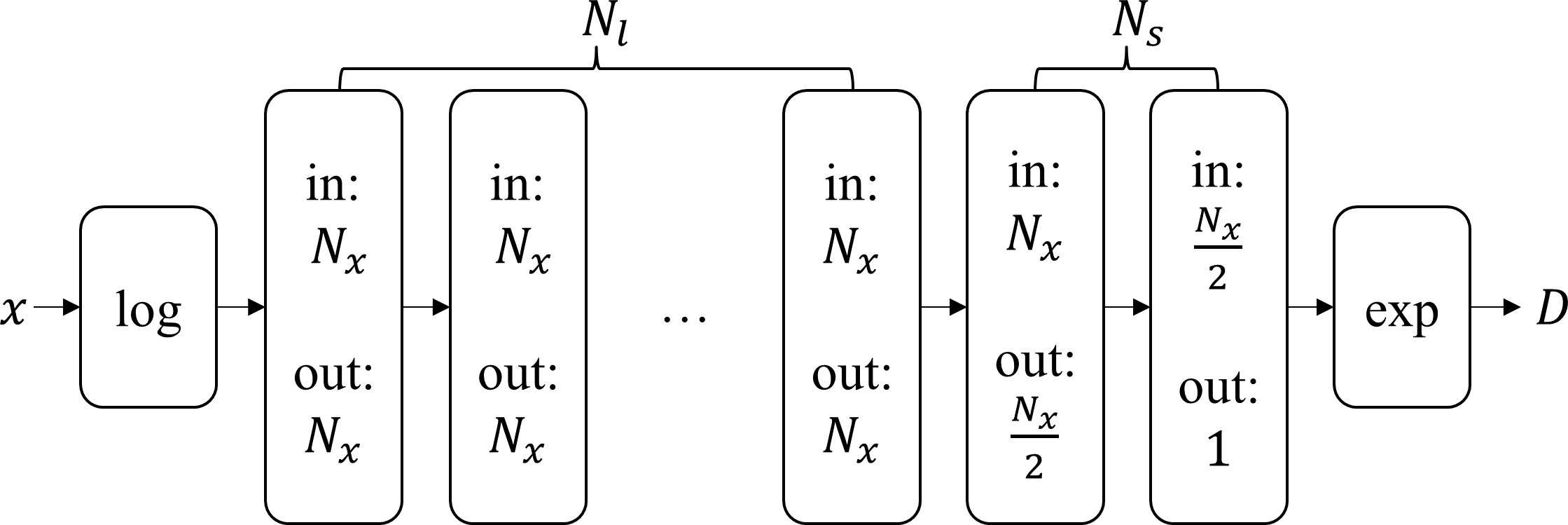}
  \caption{Network architecture: \# of linear layers is $N_l+N_s$.}
  \label{fig:net_arch} 
\end{figure}

Our network architecture is shown in Figure~\ref{fig:net_arch}. 
There are $N_l + N_s$ linear layers in total. The activation function for every layer is LeakyRelu~\cite{leakyrelu}. 
The first $N_l$ linear layers compose a set of more complex features than the raw features.
This is followed by a "summarizer" component with $N_s=2$ linear layers that gradually summarize features in $N_x$ dimensions to form the final one dimensional prediction. We can control the capacity and complexity of our model by $N_l$. We set $N_l=5$ by default and test the sensitivity to $N_l$ in Section~\ref{ssec:sensitivity}.
Since the architecture is very simple, the model inference time is at the scale of microseconds \revise{on CPU} in our experiments. 

There is one additional challenge for model learning: The magnitude of different features in the feature set $x$ can be at different scale. For example, the population size can be as large as $10^7$ while the sample NDV $d$ can be as small as $1$. This makes it very difficult to learn the model parameters~\cite{juszczak2002feature,ioffe2015batch}. The common practice to resolve this problem is to perform normalization~\cite{juszczak2002feature, ioffe2015batch}. For example in z-normalization each feature $z$ is normalized by the mean $\mu$ and standard deviation $\sigma$ of the feature: $z' = \frac{z-\mu}{\sigma}$. The underlying assumption of this practice is that training data and test data are drawn from the same distribution, so that the mean and standard deviation in the test set will be equal to that of the training set. When each test data point comes, we can use the mean and standard deviation in the training set to normalize it. However, in our case, we do not assume the training set and test set to share the same distribution. In fact, our training set is synthetically generated, so it can be very different from the real-world test datasets.

We take the logarithm of features (\eg, sample profile $f_i$'s and inverse sampling rate $1/r$) before the first layer, after adding a small constant to each feature to avoid logarithm of zero.
Since the network now operates at log scale, we take exponential on the output of the last linear layer as the final output. Specifically, let $D_{\text{log}}$ denote the output of the last linear layer, and the final estimation is $D = e^{D_{\text{log}}}$.
%
%
\revise{
There are three reasons for taking the logarithm of features in the model. First, our learning objective is to reduce the ratio error defined in \eqref{equ:errdef}, which is translated into a loss function as the squared difference between $\log D$ and $\log{\hat D}$ in \eqref{equ:loss}. Secondly, taking logarithm makes NDV and different features at the same scale, and thus easier to train the model (otherwise, the training data points with big NDVs dominate the loss). Thirdly, taking logarithm explicitly introduces non-linearity to the model, so it can be more expressive to approximate nonlinear functions, and helps the learned model to generalize better outside the support region of training data~\cite{hines1996logarithmic} to approximate unbounded functions.
}
%

\subsection{Model Regularization}
\label{sec:learning:reg}
We now introduce a regularization method tailed for our task by considering negative results on sample-based NDV estimation. Note that the negative result \cite{charikar2000towards} introduced in \csec\ref{sec:pre:ndv:negative} holds for learned estimator as well, and it says that, any learned estimator $\est \in \estclass$ with sample or sample profile as the feature has a ratio error at least in the order of $\bigomegainline{\sqrt{N/n}}$ {\em in the worst case} (for some dataset), where $N$ is the population size and $n$ is the sample size. Since the sample profile $f$ is an input feature to the learned estimator, we first try to derive an instance-wise negative result by answering the question how large the error could be after observing $f$.

\stitle{An instance-wise negative result and its implication.}
The idea of our instance-wise negative result generalizes the one of a ``global'' negative result \cite{charikar2000towards}. Consider two columns (multi-sets) $\col_1 = \col_0 \cup \Delta_1$ and $\col_2 = \col_0 \cup \Delta_2$ sharing the common subset $\col_0$, and the difference ($\Delta_1$ v.s. $\Delta_2$) makes their NDVs $D_1$ and $D_2$ differ significantly from each other. We draw two samples from $\col_1$ and $\col_2$ with sample profiles $f_1$ and $f_2$ observed, respectively.
As long as $\col_0$ is large enough, we can show that, with high probability, both the two samples contain values only in $\col_0$, and thus $f_1$ and $f_2$ follow the same distribution which are indistinguishable. More formally:
\begin{proposition} \label{prop:hardness}
For any size-$n$ sample with profile $f$ and NDV $d$, there exist two size-$N$ columns $\col_1$ and $\col_2$ with NDVs $D_1$ and $D_2$, respectively, such that with probability at least $\gamma$ we cannot distinguish whether the observed $f$ is generated from $\col_1$ or from $\col_2$, with
\begin{equation}\label{equ:hardinstance}
D_1 = \left\lfloor \frac{N-n}{4n} \left(\ln(\frac{1}{\gamma}) - \frac{2}{e^c}\right) \right\rfloor + d \hbox{~~and~~}
D_2 = d
\end{equation}
for $\gamma \geq e^{-4n-2e^{-c}}$ and $n \geq d(\ln d + c)$.
\end{proposition}
\if\TR 1
We provide the proof of \cprop\ref{prop:hardness} in Appendix~\ref{ssec:proof_prop_1}.
\else
Proof of \cprop\ref{prop:hardness} can be found in our technical
report~\cite{url:technical_report}.
\fi
\cprop\ref{prop:hardness} suggests that, no matter how well an estimator $\est$ is trained under the loss function $\hat R_A$ in \eqref{equ:totalloss}, it is inherently hard to estimate NDV using the model $\est(f,N)$. 
Consider the two columns $\col_1$ and $\col_2$ with NDVs $D_1$ and $D_2$ constructed in \cprop\ref{prop:hardness}. Samples with profiles $f_1$ and $f_2$ are then drawn from $\col_1$ and $\col_2$, respectively. \cprop\ref{prop:hardness} says that, with high probability, $f_1$ and $f_2$ follow the same distribution, and thus the expected output of $\est(f_1, N)$ should be the same as the one of $\est(f_2, N)$, considering the randomness in drawing samples. This property of $\est$ contradicts to the fact that the true NDVs $D_1$ and $D_2$ in \eqref{equ:hardinstance} differ significantly.

Following the above discussion, we can derive an instance-wise lower bound of the ratio error  by forcing $\est(\cdot, N)$ to output $\sqrt{D_1 D_2}$.
\begin{manualtheorem}{1}
\label{theorem:b2}
For any size-$n$ sample with profile $f$ and sample NDV $d$ observed (where $n \geq d(\ln d + c)$), with probability $\gamma \geq e^{-4n-2e^{-c}}$, there exist a choice of column with NDV $D$ such that any estimation of $D$ based on $f$, \ie, $\est(f, N)$, has ratio error at least
\begin{equation}\label{eq:b2}
\err(\est(f, N) ,D) \geq \sqrt{\frac{\left\lfloor \frac{N-n}{4n} \left(\ln(\frac{1}{\gamma}) - \frac{2}{e^c}\right) \right\rfloor + d}{d}} \triangleq b(d,n).
\end{equation}
%
\end{manualtheorem}

\if\TR 1
We defer the poof of Theorem~\ref{theorem:b2} to Appendix~\ref{ssec:proof_b2}.
\else
Proof of Theorem~\ref{theorem:b2} can be found in our technical
report~\cite{url:technical_report}.
\fi

\stitle{Regularization for a worst-case optimal estimator.}
When training an estimator $\est$ under the loss function $\hat R_A$ in \eqref{equ:totalloss}, consider two training data points $p_1 = ((f, N), D_1)$ and $p_2 = ((f, N), D_2)$ where $D_1$ and $D_2$ as in \eqref{equ:hardinstance} are NDVs of the two columns $\col_1$ and $\col_2$ constructed in \cprop\ref{prop:hardness}. If $p_1$ is in the training data $A$ but $p_2$ is not, $\est$ tends to predict $D_1$ after training; if $p_2$ is in $A$ but $p_1$ is not, a trained $\est$ tends to predict $D_2$. In both cases, the worst-case ratio error for estimation could be as large as $D_1/D_2$.

Our goal of regularization here is to push a trained model $\est$ towards an {\em instance-wise worst-case optimal estimator} $\est^*$ whose ratio error matches an {\em instance-wise lower bound} $b(d,n)$ as in \eqref{eq:b2}.
To this end, consider a loss function $\hat R_{A,\est^*}(\est)$ which aims to minimize the distance between $\est$ and $\est^*$ in prediction:
\begin{equation} \label{equ:totallossreg}
\hat R_{A,\est^*}(\est) = \frac{1}{|A|}\sum_{((f, N), D) \in A} |L(\est(f, N), D) -  L(\est^*(f, N), D)|.
\end{equation}
%
Intuitively, with this loss function in \eqref{equ:totallossreg}, we want our model $\est$ to be trained towards the ``optimal'' estimator $\est^*$. In particular, if $\hat R_{A,\est^*}(\est) = 0$, $\est$ behaves exactly the same as $\est^*$ on the training set $A$ with $L(\est(f, N), D) =  L(\est^*(f, N), D)$ for each $((f, N), D) \in A$.
Comparing to the loss $\hat R_A(\est)$ in \eqref{equ:totalloss}, $\hat R_{A,\est^*}(\est)$ in \eqref{equ:totallossreg} allows the model $\est$ to have a higher error (only as good as $L(\est^*(f, N), D)$) on the training set, but aims to prevent overfitting.

\stitle{Loss function with regularization.}
Recall the loss $L(\cdot, D)$ in \eqref{equ:loss} used by our model. If $\est^*$'s ratio error matches the {\em instance-wise lower bound} $b(d,n)$ in \eqref{eq:b2}, we have $L(\est^*(f, N), D) = (\log b(d,n))^2$. We also apply L2 regularization on model parameters $W$ to encourage $W$ to be small and sparse for better generalization~\cite{bishop2006pattern}. Putting them together, we use the following loss function in training:
\begin{equation} \label{equ:losswithreg}
\!\!\hat R^*_A(\est) = \frac{1}{|A|}\!\!\sum_{((f, N), D) \in A}\!\!\!\!\!\!\!\! |L(\est(f, N), D) -  (\log b(d,n))^2| + \lambda \|W\|_2.
\end{equation}
Note that $b(d,n)$ is defined only when $n \geq d(\ln d + c)$ according to \cthm\ref{theorem:b2}. For $n < d(\ln d + c)$ (sample NDV is close to sample size), we define $b(d,n) \triangleq 1$. In our implementation, $c$ is set to be $10$ in \eqref{eq:b2} such that the $2/e^c$ term in $b(d,n)$ is negligible.

There are two hyperparameters here. i) $\gamma$ for $b(d,n)$ defined in \eqref{eq:b2} controls how confident this lower bound is. We set $\gamma=0.6$ by default to have a medium level of confidence.
%
ii) $\lambda$ controls the strength of the L2 regularization. 
%
%
%
\if\TR 1
It is typically selected through a validation set that is assumed to be similar to the test set. However, we do not have access to such a validation/test set.
Intuitively, while it is better to have sufficiently small training loss, we want to set $\lambda$ to be as large as possible. Based on this intuition, we select $\lambda$ such that any further increase of $\lambda$ will increase the loss significantly after training. We vary $\lambda$ at different scales, \ie, $0$,$10^{-4}$, $10^{-3}$, $10^{-2}$, $10^{-1}$, $1$, $10^{1}$. \cfig\ref{fig:d_loss_vs_wd} shows the training loss under different $\lambda$. Training loss increases relatively significantly when $\lambda$ increases from $10^{-1}$ to $10^0$. Therefore, we set $\lambda=10^{-1}$ by default in our trained model.

\begin{figure}[htb!]
\Description{}
  \centering
  \includegraphics[width=0.6\linewidth]{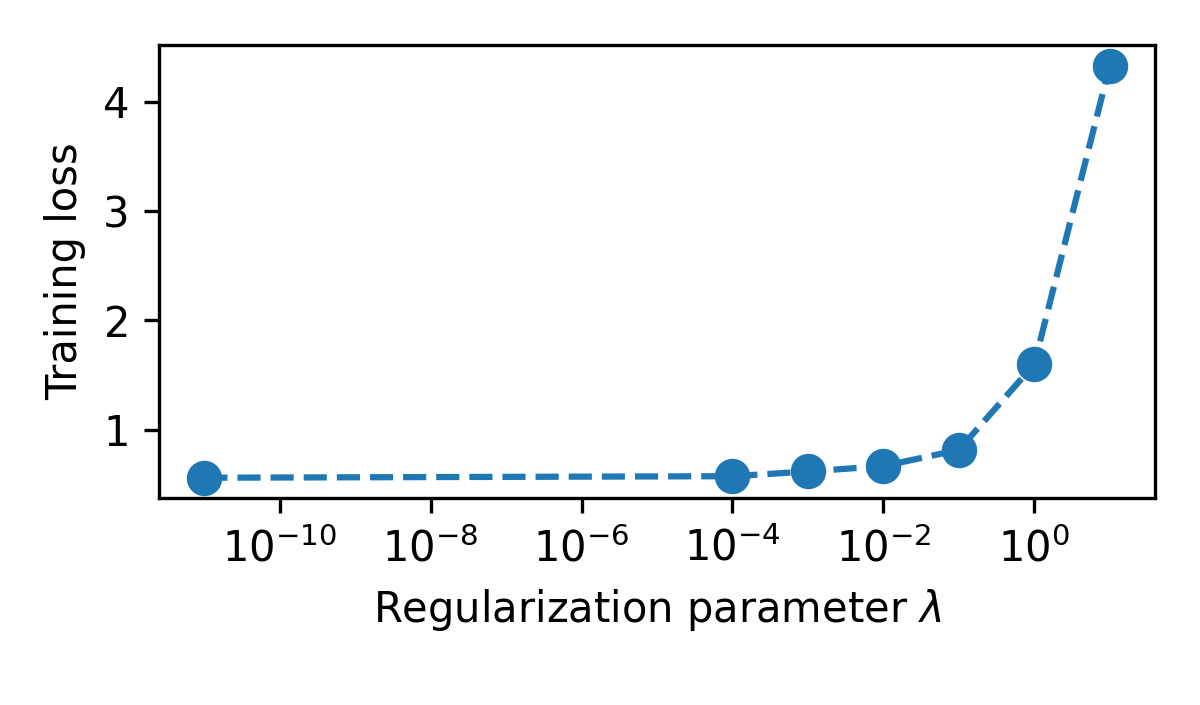}
  \caption{Training loss vs L2 regularization parameter $\lambda$.}
  \label{fig:d_loss_vs_wd} 
\end{figure}

\else
We set $\lambda=10^{-1}$ by default in our learned estimator after a tuning process based on training loss.
Details about the tuning process can be found in our technical report \cite{url:technical_report}.
\fi
It is important to note that, when choosing $\gamma$ and $\lambda$, we use absolutely no knowledge about the test datasets.
The robustness of our model to different choices of $\gamma$ and $\lambda$ is evaluated in \csec\ref{ssec:sensitivity}.

%% file: paper-experiments.tex
\section{Usage and Deployment}
\label{sec:use}
We extract the trained weights of our model in ``\texttt{model\_paras.npy}'' and implement a numpy version of the model to have minimum dependency on other libraries. We provide the trained model, \ie, our learned NDV estimator in~\cite{url:est}. The usage of it is as simple as a statistical estimator such as GEE and users can easily estimate population NDV by providing either a sample or a sample profile (drawn from a column or query results), as shown below:
\begin{figure}[htb!]
  \centering
  \vspace{-3mm}
  \includegraphics[width=\linewidth]{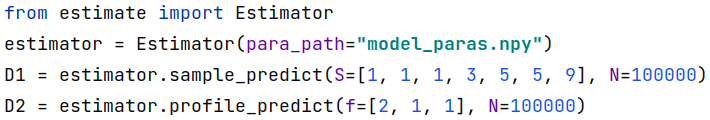}
  \label{fig:example_usage} 
  \vspace{-8mm}
\end{figure}

Our model does not require any re-training for new workloads and the inference time is at the scale of microseconds \revise{on CPU} as our model structure is very simple.
\revise{Specifically, in our experiments, 
inference on a single profile has a running time of $240\times 10^{-6}$ seconds.}
Our trained estimator can be easily plugged into any existing systems. For example, we have plugged our trained model to the cloud big data processing platform MaxCompute~\cite{maxcompute} at Alibaba. 

\section{Experiments}
\label{sec:exp}
We conduct experiments to evaluate the efficacy of our proposed method and compare it with baselines along three dimensions:
\begin{itemize}[leftmargin=*]
    \item \textit{Performance.} How accurate is our learned estimator?
    \item \textit{Ablation study.} How does our way of generating training data and performing model training contribute to the final performance?
    \item \textit{Sensitivity analysis.} How sensitive is the performance of our model to different hyperparameters?
\end{itemize}
\subsection{Experiment Setup}
\noindent\textbf{Hardware.} All our experiments were performed on a machine with a 2.50GHz Intel(R) Xeon(R) Platinum 8163 CPU, a GeForce RTX 2080 Ti GPU and 376GB 2666MHz RAM.

\noindent\textbf{Datasets.}  We conduct experiments on nine real-world datasets from diverse domains. Note our method does not use any of them for any training or hyperparameter tuning. The datasets are only used for evaluation after our model is trained on synthetic data.
\begin{enumerate}[wide,labelindent=0pt,label=\arabic*)]
\item \emph{Kasandr}~\cite{sidana2017kasandr,ting2019approximate}: Behavior records of customers in e-Commerce advertising with 15.8M rows and 7 columns.
\item \emph{Airline}~\cite{bts,airline_data}: Summary statistics of airline departures from 1987 to 2013 with 10.0M rows and 10 columns.
\item \emph{DMV}~\cite{data_dmv,kiefer2017estimating}: Data from Department
of Motor Vehicles, containing information about cars, their owners and accidents. There are 11.7M rows and 20 columns.
\item \emph{Campaign}~\cite{Campaign_finance}: Information of individual contributions to election campaigns. There are 3.3M rows and 21 columns.
\item \emph{SSB}~\cite{o2007star}: The star schema benchmark. We use the fact table with a scaling factor of 50, resulting in 300M rows and 17 columns.
\item \emph{NCVR}~\cite{nvcr}: North Carolina voter registration data with 8.3M rows and 71 columns. 
\item \emph{Product}: Private dataset with information of product items on an online-shopping website. There are 5.2M rows and 25 columns.
\item \emph{Inventory}: Private dataset of inventory statistics. There are 8.8M rows and 19 columns
\item \emph{Logistics}: Private dataset of logistics information of shipping orders. There are 8.6M rows and 28 columns.
\end{enumerate}

\noindent\textbf{Methods evaluated.} We compare our method to nine methods:
\begin{enumerate}[wide,labelindent=0pt,label=\arabic*)]
\item \emph{GEE}~\cite{charikar2000towards}: This method is constructed by using geometric mean to balance the two extreme bounds of NDV. It is proved to match a theoretical lower bound of ratio error within a constant factor.
\item \emph{HYBGEE}~\cite{charikar2000towards}: This is a hybrid estimator using GEE for high-skew data and using the smoothed jackknife estimator for low-skew data. We refer to Pydistinct~\cite{pydistinct} to test for high or low skewness. 
\item \emph{HYBSKEW}~\cite{haas1995sampling}: This is a hybrid estimator using shlosser for high-skew data and using the smoothed jackknife estimator for low-skew data. We refer to the implementation in Pydistinct~\cite{pydistinct}.
\item \emph{AE}~\cite{charikar2000towards}: This was proposed to be a more principled version of HYBGEE with smooth transition from low-skew data to high-skew data. This method requires numerically finding the root of a non-linear equation. We use the classic Brent’s method~\cite{brentq,brent1973algorithms}.
\item \emph{Chao}~\cite{chao1984nonparametric}: This method is derived by approximating the coverage as $1-f_1/n$ and assuming the population size is infinity. It predicts NDV to be infinite when $f_2=0$, so we invoke GEE in this case.

\item \emph{Chao-Lee}~\cite{chao1992estimating}: 
This method adds a correction term to the coverage estimation in Chao to handle skew in data~\cite{haas1995sampling}.
\item \emph{Shlosser}~\cite{shlosser1981estimation}: This method is derived based on a skewness assumption: $E(f_i)/E(f_1) \approx F_i/F_1$. The method performs well when each distinct value appear approximately one time on average~\cite{haas1995sampling}.

\item \emph{APML}~\cite{pavlichin2019approximate}: This method analytically approximates the profile maximum likelihood estimation for population profile $F^{\text{PML}}$ (solution to \eqref{equ:fpml}), and estimate NDV as $\sum_i F^{\text{PML}}_i$. We use the implementation provided in the original paper. Note this method doesn't assume population size $N$ is given. For a fair comparison, when this method predicts NDV to be greater than $N$ (in this case ratio error can be extremely big), we replace its prediction with the best performing baseline GEE's prediction. 
\item \emph{Lower Bound (LB)}: This method is to demonstrate the possible gain of accessing some columns of the real-world test set at training phase. For this method, we split all columns in the the nine real-world datasets by 4:1:5 as training, validation and test set. We fine tune our learned model (trained on synthetic data) on the training set and use the validation set to perform early stopping to prevent over-fitting. We evaluate the fine-tuned model on the remaining test set. This provides an empirical lower bound for the error of any workload agnostic estimators.
\end{enumerate}

\begin{table}[t]
\setlength{\tabcolsep}{0.1em}
\renewcommand{\arraystretch}{0.9}
\caption{\revise{Inference complexity (1st line) and \# arithmetic operations (2nd line) needed in implementation}}
\scalebox{0.98}{\footnotesize
\begin{tabular}{|l|l|l|l|l|l|l|l|l|l|}
\hline
GEE & \makecell{HYB\\GEE} & \makecell{HYB\\SKEW} & AE & Chao & \makecell{Chao\\-Lee} & \makecell{Shlo\\sser} & APML & Our \\ \hline
$\bigohinline{1}$ & $\bigohinline{n}$ & $\bigohinline{n}$ & - & $\bigohinline{1}$ & $\bigohinline{|f|}$ & $\bigohinline{|f|}$ & $\bigohinline{n}$ & $\bigohinline{1}$ \\ \hline
$5$ & $5n$  & $5n$ & - & 4 & $4|f|$ & $7|f|+2$ & $n+\sqrt{n}\log n$ & $300+51200$ \\ \hline
\end{tabular}
}
\label{tbl:inference_complexity}
\end{table}

\noindent\revise{\noindent\textbf{Inference cost.} For all the methods evaluated, the inference cost, \ie, {\em processing time of estimating NDV per column}, is usually dominated by the cost of scanning the sample with size $n$ to obtain sample profile $f$ and sample NDV $d$. {\em Suppose $f$ and $d$ are already obtained}, \ctab\ref{tbl:inference_complexity} summarizes the complexity and the number of arithmetic operations needed in the {\em remaining steps} for inference in different methods. Here, let $|f| = \max\{j \mid f_j>0\}$ denote the length of $f$. GEE, Chao, and our method need another $\bigohinline{1}$ arithmetic operations to derive the NDV estimation, while Chao-Lee and Shlosser need another $\bigohinline{|f|}$ arithmetic operations and HYBGEE, HYBSKEW, and APML need another $\bigohinline{n}$ operations. Our method needs about 300 arithmetic operations for feature engineering and 51200 arithmetic operations for a forward pass in the neural network. AE needs to solve a nonlinear equation numerically.}

\noindent \textbf{Setup for various estimators.} We evaluate all methods with sampling rates varying from $10^{-4}$ to $10^{-2}$: $10^{-4}$, $2\times 10^{-4}$, $5\times 10^{-4}$, $10^{-3}$, $2\times 10^{-3}$, $5\times 10^{-3}$, and $10^{-2}$. To deal with randomness in e.g. sampling, we run ten times and report the averaged results. We implement our model with pytorch and use Adam~\cite{kingma2014adam} as optimizer and perform training with skorch~\cite{skorch}. We implement L2 regularization with its equivalent form - weight decay~\cite{metacademy_wd, van2017l2}. 
For our method, the hyperparameters include: the number of elements in sample profile that are used as features $m$, the number of linear layers in our network architecture $N_l$, learning rate during training $lr$, logarithm of the smallest sampling rate $B'$, logarithm of the largest population size in training data $B$, number of synthetic training data used $N_A$, the confidence level $\gamma$ of our derived lower bound in \eqref{eq:b2}, and the L2 regularization parameter $\lambda$. We have explained we set $N_l=5$ in Section~\ref{sssec:model_struc}, $B'=-4$ in Section~\ref{sssec:sample_profile_gen}, $B=9$ in Section~\ref{sssec:pop_profile_gen},
$\gamma=0.6$ and $\lambda=0.1$ in Section~\ref{sec:learning:reg}. 
For the remaining parameters: We set $m=100$ so the total number of features we use is $N_x=106$, and we test the sensitivity to $m$ in Section~\ref{ssec:sensitivity}.
We choose a learning rate that has the smallest training loss and also converges in a reasonable time: $lr=0.0003$.
We generate $N_A = 0.72 \times 10^6$ training data points, which takes about two hours running in parallel on our machine and we further vary the amount of training data used in Section~\ref{ssec:sensitivity}.
For sanity, we drop the data points with population size $N<10^4$ as the sample may contain zero data points because sampling rate can be as small as $10^{-4}$.

\noindent \textbf{Performance metric.} We used the widely used ratio error as our performance metric, which is defined in \eqref{equ:errdef}.

\subsection{Performance}
\stitle{Overall performance.}
The overall performance of all methods is shown in Table~\ref{tbl:overall_performance}. Our method achieves the lowest ratio error on seven out of nine datasets and has comparable error to the best baseline methods on the other two datasets. 
Our averaged ratio error is very close to the empirical lower-bound error.  
Overall, most baseline methods fail significantly on at least one dataset, which could be caused by the violation of the assumptions they make.  GEE is the best performing baseline. Although HYBGEE and AE were both designed to improve over GEE, their performance is very sensitive to choice of dataset, because HYBGEE involves estimating skewness of population which can be difficult on some datasets and AE requires numerically solving a non-linear equation which can be brittle in some cases. APML also doesn't work well. One reason is the analytical approximation in APML can introduce a big error; Another reason is that APML is designed to estimate population profile $F$ and is not tailored for estimating NDV.

\begin{figure}[htb!]
  \centering
  \includegraphics[width=\linewidth, height=6.5cm]{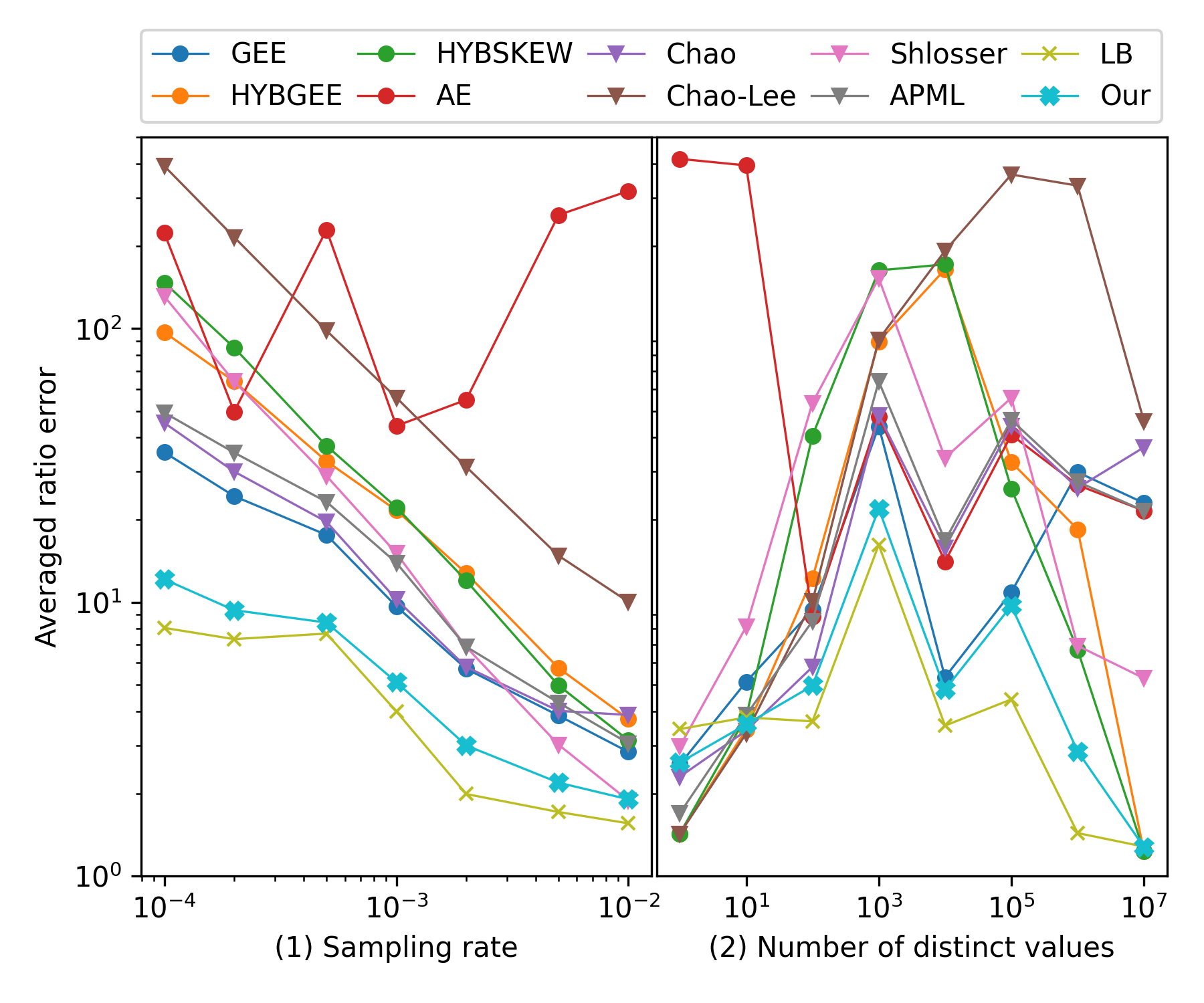}
  \caption{Averaged ratio error over all datasets for different (1) samping rate and (2) NDV.}
  \label{fig:err_vs_rate_ndv} 
\end{figure}

\stitle{Ratio error under different sampling rate.}
We plot the averaged ratio error under different sampling rate in Figure~\ref{fig:err_vs_rate_ndv}(1). Our method has the lowest error and is close to the empirical lower-bound under all sampling rates. In addition, as sampling rate decreases, the advantage of our method over other methods increases significantly. Shlosser has comparable performance with our method at a high sampling rate $10^{-2}$, but its performance decreases rapidly as sampling rate decreases. The error of AE is irregular with respect to sampling rate because error of AE depends on the accuracy of the root finding of a non-linear equation, which is brittle in practice.

\begin{table}[t]
\setlength{\tabcolsep}{0.1em}
\renewcommand{\arraystretch}{0.9}
\caption{Ratio error for all methods averaged over all columns in each dataset and over all sampling rates. The error for the empirical lower bound is denoted in grey.}
\scalebox{0.98}{
\begin{tabular}{|l|l|l|l|l|l|l|l|l|l|l|}
\hline
           & GEE  & \makecell{HYB\\GEE} & \makecell{HYB\\SKEW} & AE           & Chao         & \makecell{Chao\\-Lee} & \makecell{Shlo\\sser} & APML & \textcolor{gray}{LB} & Our           \\ \hline
Kasandr    & 3.4  & 7.9  & 8.2   & 4.6          & 4.7          & 9.9      & 19.2     &    5.1  &    \textcolor{gray}{2.2}        & \textbf{2.4}  \\ \hline
Airline    & 4.1  & 1.8  & 1.8   & \textbf{1.4} & 1.6          & 1.8      & 70.0     &  2.0    &           \textcolor{gray}{2.5} & 1.9           \\ \hline
DMV        & 5.2  & 3.5  & 17.5  & 7.7          & 8.8          & 13.8     & 23.5     &    7.6  &           \textcolor{gray}{3.2} & \textbf{2.7}  \\ \hline
Campaign   & 7.3  & 6.5  & 8.4   & 291.8        & 13.4         & 48.0     & 21.9     &      13.2&           \textcolor{gray}{2.9} & \textbf{3.9}  \\ \hline
SSB        & 5.2  & 1.2  & 1.2   & \textbf{1.1} & \textbf{1.1} & 1.2      & 173.1    & 1.3     &            \textcolor{gray}{1.5}& 2.0           \\ \hline
NCVR       & 12.2 & 31.0 & 56.9  & 150.2        & 13.2         & 58.6     & 42.6     &  16.9    &           \textcolor{gray}{5.6} & \textbf{6.4}  \\ \hline
Product    & 36.3 & 60.6 & 58.7  & 46.1         & 46.9         & 250.6    & 30.5     &  54.8    &           \textcolor{gray}{9.2} & \textbf{14.6} \\ \hline
Inventory  & 17.8 & 23.5 & 18.4  & 75.1         & 24.8         & 252.9    & 11.6     &   26.7   &            \textcolor{gray}{4.5}& \textbf{7.8}  \\ \hline
Logistics & 17.1 & 93.1 & 100.0 & 552.0        & 15.5         & 275.1    & 19.5     &   16.7   &            \textcolor{gray}{3.6}& \textbf{3.5}  \\ \hline
Average    & 12.1 & 25.5 & 30.1  & 125.6        & 14.5         & 101.3    & 45.7     &  16.0    &           \textcolor{gray}{3.9} & \textbf{5.0}  \\ \hline
\end{tabular}
}
\label{tbl:overall_performance}
\end{table}

\stitle{Ratio error under different NDV.}
We plot the averaged ratio error under different number of distinct values in Figure~\ref{fig:err_vs_rate_ndv}(2). Our method has the lowest error under all range of number of distinct values except the extremely small region, \ie, NDV $\approx 1$.

\revise{Data distributions with high skew (small NDV) and low skew (large NDV) are two relatively ``easy'' scenarios for NDV estimations. 
Intuitively, when NDV is extremely small, it can be estimated as NDV in the sample; when NDV is extremely large (close to population size), NDV can be estimated as $\frac{{\rm sample~NDV}}{{\rm sampling~rate}}$.
For example, as is analyzed in \cite{charikar2000towards} and demonstrated above in Figure~\ref{fig:err_vs_rate_ndv}(2), GEE and Shlosser are known to perform well for data with high skew (NDV $<10$); some other estimator, \eg, HYBSKEW and HYBGEE, combines the results of two estimators (one performs well for low skew and the other for high skew)--one of the two is selected depending on a test designed to measure the skew of the data.
Therefore, the general trends for most estimators observed in Figure~\ref{fig:err_vs_rate_ndv}(2) are similar: they perform well for small NDV, and as NDV increases, the error first increases and then decreases for large NDVs.
Given this general trend, however, the turning points for different methods are different, depending on their concrete forms and data distributions. For example, the turning point of Chao-Lee is $10^5$; the turning point of HYBSKEW/HYBGEE is about $10^4$, and their performance also largely depends on the skewness test result on the sample data, which may not be stable; the turning point of Shlosser is $10^3$.
The performance of AE is irregular and largely depends on data distributions, as it considers only estimators of the form $d + K f_1$ \cite{charikar2000towards}, and tries to estimate $K$ from the sample data.}

%

%

\begin{figure}[t]
  \centering
  \includegraphics[width=0.8\linewidth]{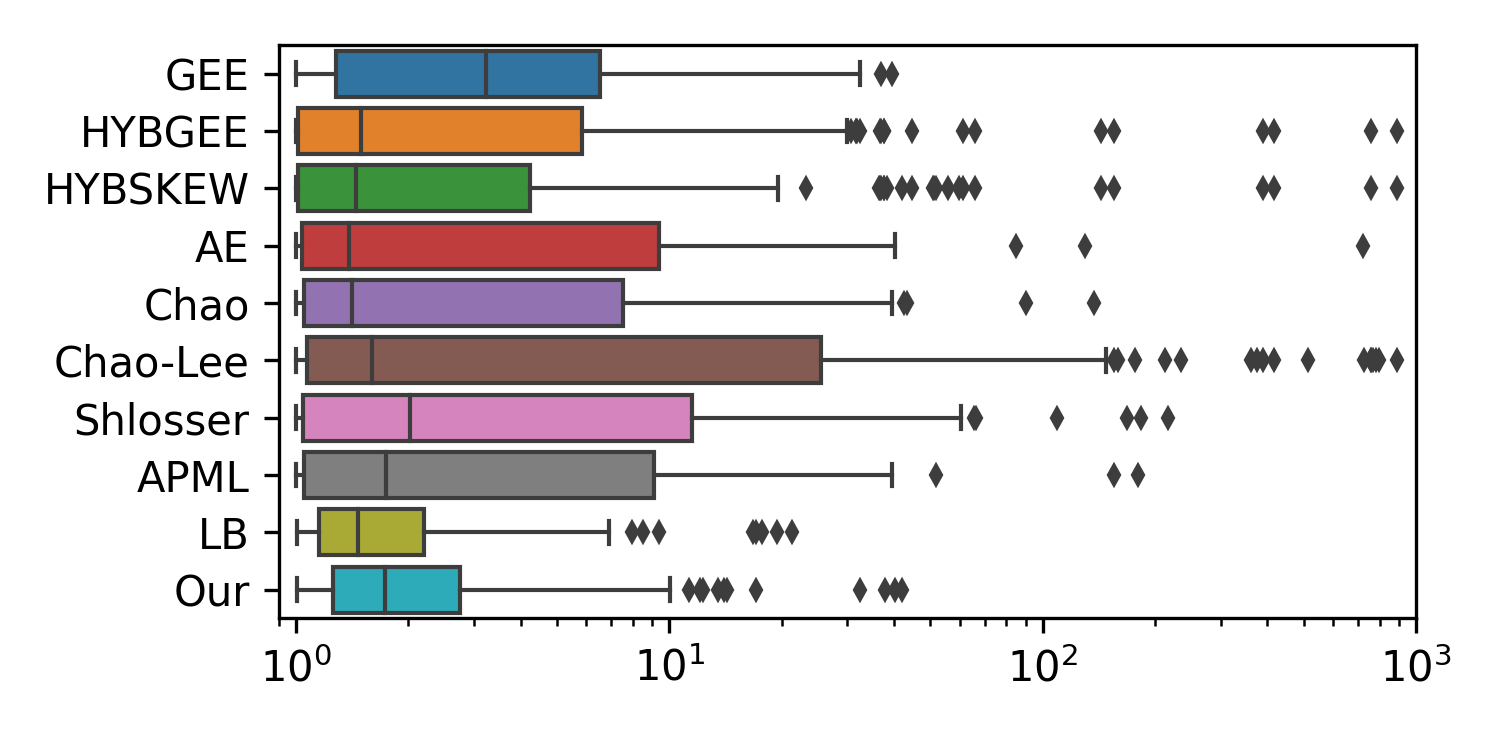}
  \caption{Ratio error on all columns in all datasets when sampling rate $r=10^{-3}$. In each sub-boxplot for each method, the five vertical lines denote minimum, $25$th percentile, median, $75$th percentile, and maximum respectively; the colored box contains $50\%$ of the data points; the diamond symbols on the right side of the maximum line are outliers.}
  \label{fig:err_boxplot} 
\end{figure}

\stitle{Ratio error distribution.}
To show the distribution and the worst case of ratio error, we plot the boxplot~\cite{box_plot} of ratio error in Figure~\ref{fig:err_boxplot} when sampling rate $r=10^{-3}$. Overall, our method has the smallest error in most cases and also has the smallest worst case error. In $75\%$ of cases, the ratio error of our method is smaller than $3$, much better than all baselines. The worst case ratio error of our method is almost the same as the baseline GEE that has theoretical worst case ratio error guarantee. In $75\%$ of cases, the two hybrid methods HYBGEE and HYBSKEW are on average better than GEE but they are not stable and their worst case error can be as large as $1000$ causing their averaged ratio error to be greater than GEE in Table~\ref{tbl:overall_performance}.

\subsection{Ablation Study}
We perform ablation study to show the contribution of each component in our method and compare them to existing or straightforward solutions. Table~\ref{tbl:ablation} shows the results. Column "Full" denotes our method with full set of components. We ablate our method in three aspects, training data generation, model architecture and objective function for training. The symbol "-" denotes dropping the component that it precedes. 

\noindent\textbf{Training data generation.} As shown in column "-diversify" in Table~\ref{tbl:ablation}, when removing the diversify component proposed in Section~\ref{sssec:diversify}, ratio error increases a bit, though the error is still significantly lower than the best performing baseline in Table~\ref{tbl:overall_performance}. This verifies our intuition in Section~\ref{sssec:diversify}: increasing the support region of the profiles is helpful to the generalization ability of our model. We also compare to four heuristic method of generating training data. "uni" denotes generating each $F_i$ of the population profile $F$ by a uniform distribution. "pl" denotes generating $F$ so that the relationship between $F_i$ and $i$ is power law. "rw" denotes generating $F$ by random walk, i.e. $F_{i+1}=F_{i}+s$ where $s$ is a random step size. "mix" denotes mixing all training data generated by the three heuristic methods. For the three heuristic methods, we also ensure that we introduce enough randomness by using a set of different random hyperparameters (e.g. length of $F$, range of uniform distribution, parameters in power law distribution, and step size in random walk) for each data point. The results in Table~\ref{tbl:ablation} show that our way of generating training data outperforms the heuristic methods significantly. When mixing the three types of heuristically generated data together, the averaged error is smaller than using each type along because mixing all data increases training data diversity and helps the model to generalize better.

\noindent\textbf{Model architecture.} We drop the logarithm layer in our model architecture (Figure~\ref{fig:net_arch}). As shown in column "-log" in Table~\ref{tbl:ablation}, the ratio error increases dramatically. This validates our choice to take logarithm and its advantages as discussed in \csec\ref{sssec:model_struc}.
%

\noindent\textbf{Training objective.} Column "-b" in Table~\ref{tbl:ablation} denotes removing our proposed regularization for a worst-case optimal estimator in \eqref{equ:totallossreg}.
Overall, contrasting the column "Full" to the column "-b", we can see adding the proposed regularization makes the overall ratio error decrease by $0.9$, closing the gap to the empirical lower-bound by $45\%$. 
\revise{
This regularization pushes the model to a worst-case optimal estimator, and make it more robust (for datasets where it is not the best, it is close to the best-performing estimator). With this regularization, the ratio error on the most difficult datasets (\eg, Product, and Inventory) decreases significantly more than on other datasets.
%
%
Removing this regularization does not make the learned estimator "optimal" in specific datasets, but the error may decrease a bit on the easy datasets (\eg, Kasandr, Airline, and DMV).}
%
%
%
We further drop the L2 regularization (column "-b-L2") and error further increases. This is expected as L2 regularization encourages sparse and small model parameters that generalize better. 
Finally, we further drop the logarithm on NDV (column "-b-L2-log") to use a naive mean squared loss on NDV and the error increases to be extremely high (similar to the "-log" column under "Model arch").

\begin{table}[t]
\setlength{\tabcolsep}{0.1em}
\renewcommand{\arraystretch}{0.9}
\caption{Ablation analysis. "-" denotes removing a component. The four columns under "heuristic" replace our training data generation by four heuristic methods respectively.}
\begin{tabular}{|l|l|l|l|l|l|l|l|l|l|l|}
\hline
\multirow{3}{*}{Datasets} & \multirow{3}{*}{Full} & \multicolumn{5}{l|}{\makecell{Training data\\ generation}} & \makecell{Model\\ arch} & \multicolumn{3}{l|}{\makecell{Training\\ objective}} \\ \cline{3-11} 
 &  & \multirow{2}{*}{\makecell{-diver\\sify}} & \multicolumn{4}{l|}{heuristic} & \multirow{2}{*}{-log} & \multirow{2}{*}{-b} & \multirow{2}{*}{\makecell{-b\\-L2}} & \multirow{2}{*}{\makecell{-b-L2\\ -log}} \\ \cline{4-7}
 &  &  & uni & rw & pl & mix &  &  &  &  \\ \hline
Kasandr & 2.4 & 5.2 &40.8  & 30.8 & >1k & 50.1 & >1k & 2.3 & 3.3 &>1k  \\ \hline
Airline & 1.9 & 2.5 &3.4  & 4.7 & 464.0 & 24.9 &>1k  & 1.8 & 1.9 &>1k  \\ \hline
DMV & 2.7 & 3.4 &129.8  & 45.2 & 14.6 & 98.3 &>1k  & 2.5 & 3.8 & >1k \\ \hline
Campaign & 3.9 & 6.7 &147.9  & 176.5 & 933.1 & 127.9 &>1k  & 3.9 & 21.6 & >1k \\ \hline
SSB & 2.0 & 3.3 &19.1  & 1.2 & 6.5 & 3.1 &>1k  & 2.1 & 1.8 &>1k  \\ \hline
NCVR & 6.4 & 8.1 &83.2  & 92.9 & >1k & 113.8 &>1k  & 7.2 & 8.2 &>1k  \\ \hline
Product & 14.6 & 20.1 &417.6 & 227.3 & 52.7 & 658.5 &>1k  & 20.4 & 22.0 &>1k  \\ \hline
Inventory & 7.8 & 10.1 &386.7  & 313.4 & >1k & 109.5 &>1k  & 9.4 & 12.0 &>1k  \\ \hline
Logistics & 3.5 & 9.2 &496.7  & 390.6 & >1k & 115.5 &>1k  & 3.8 & 4.8 &>1k  \\ \hline
Average & 5.0 & 7.7 &191.2  & 142.5 & >1k & 76.9 &>1k  & 5.9 & 8.8 &>1k  \\ \hline
\end{tabular}
\label{tbl:ablation}
\end{table}

\begin{figure*}[htb!]
  \centering
  \includegraphics[width=16cm]{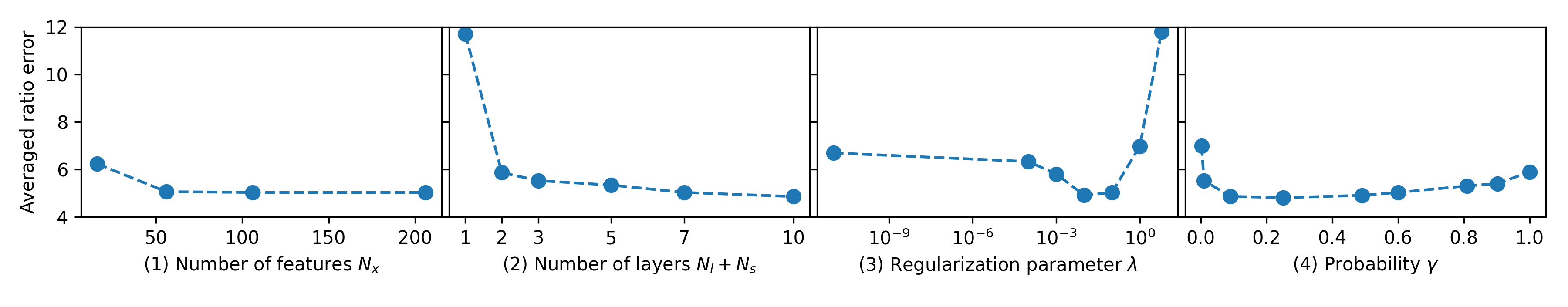}
  \caption{Averaged ratio error vs (1) \# of features $N_x$ (2) \# of layers $N_l+N_s$ (3) L2 regularization parameter $\lambda$ (4) Probability $\gamma$}
  \label{fig:err_vs} 
\end{figure*}

\subsection{Sensitivity Analysis}
\label{ssec:sensitivity}
We show the robustness of our method to the hyperparameters. 
\noindent\textbf{Sensitivity to number of features.} Our number of features is $N_x = m+6$ where $m$ is the number of elements in sample profile that used as raw features. We set $m = 100$ by default. To test the sensitivity, we vary $m$ from 10 to 200. As shown in Figure~\ref{fig:err_vs}(1), as $m$ increases from 100 to 200 ($N_x$ increases from 106 to 206), averaged ratio error doesn't change, so setting $m=100$ suffices. 

\noindent\textbf{Sensitivity to number of layers in network.} 
The number of layers in our network is $N_l + N_s$ where $N_s=2$. We vary $N_l$ from $0$ to $8$ so the number of layers varys from $2$ to $10$. 
In addtion, we evaluate one extreme case $N_l = 0$ and $N_s=1$, where the network only has one linear layers with $N_x$ input dimenions and one output demension. The averaged ratio error with respect to the number of layers is shown in Figure~\ref{fig:err_vs}(2). As the number of layer increases, averaged ratio error decreases due to the increase of model complexity. 
Although $N_l+N_s=10$ gives marginally better result, our choice with $N_l=5$ and  $N_l+N_s=7$ gives good enough result. 


\noindent\textbf{Sensitivity to L2 regularization.} Figure~\ref{fig:err_vs}(3)
shows the average ratio error under different L2 regularization parameter $\lambda$. The error is overall quite stable when $\lambda$ is in region [$0, 10^{-1}$]. Error increases significantly when $\lambda$ increases from $10^{-1}$ to 1 and increases even more dramatically when $\lambda$ increases from 1 to 10. This increasing trend in region $[10^{-1},+\infty)$ is similar to the increasing trend of the training loss\if\TR 1
in Figure~\ref{fig:d_loss_vs_wd}.
\else
\cite{url:technical_report}. 
\fi
Our heuristic method of selecting $\lambda$ based on the training loss curve at the end of Section~\ref{sec:learning:reg} is able to select $\lambda=10^{-1}$. Although the optimal $\lambda$ is at $10^{-2}$, $\lambda=10^{-1}$ is good enough. 

\noindent\textbf{Sensitivity to probability $\gamma$.} Probability $\gamma$ is the confidence level of our derived lower bound in \eqref{eq:b2}. Figure~\ref{fig:err_vs}(4)
shows the averaged ratio error when probability $\gamma$ varies in region $(0,1]$. Overall, the averaged ratio error is quite stable. The optimal value for $\gamma$ is about $0.2$. We heuristically selected $\gamma$ to be $0.6$ in Section~\ref{sec:learning:reg}. Although it is not the optimal,  $\gamma=0.6$ is good enough.

\begin{figure}[t]
	\centering
	\includegraphics[width=0.7\linewidth]{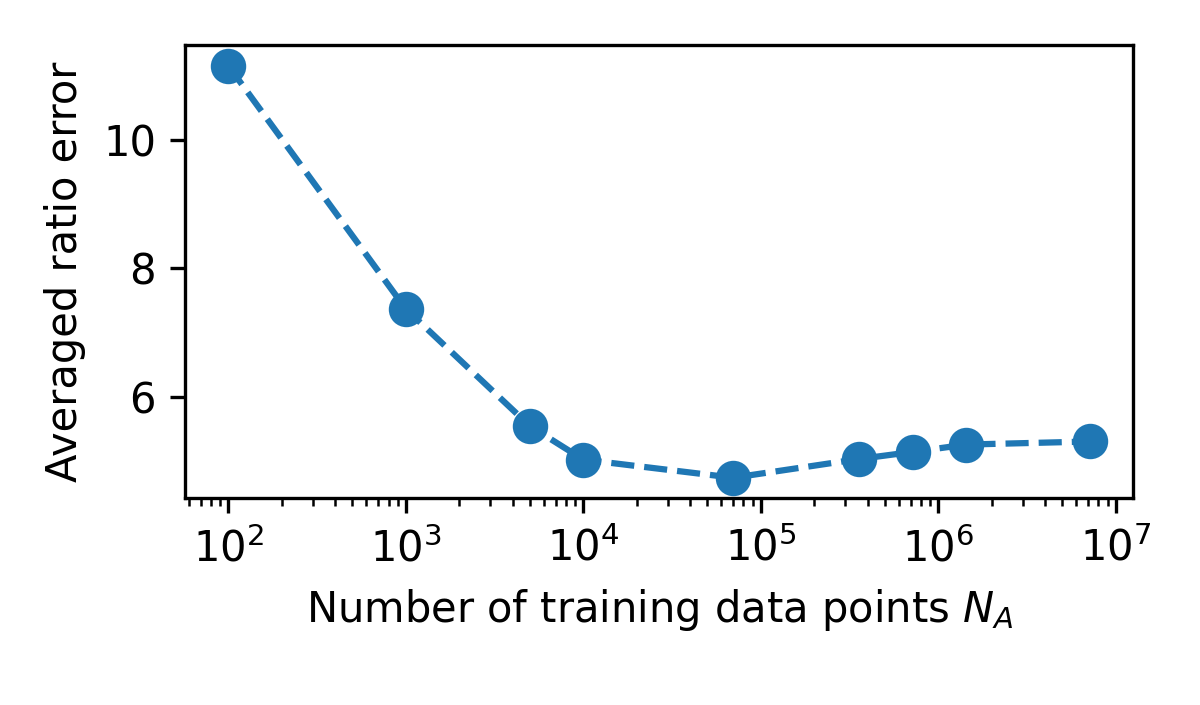}
	\caption{Averaged ratio error vs number of training data $N_A$.}
	\label{fig:err_vs_nd} 
\end{figure}

\noindent\textbf{Sensitivity to number of training data points.}
We vary the number of training data points $N_A$ from $10^2$ to $7.2 \times 10^6$. In Figure~\ref{fig:err_vs_nd}, the averaged ratio error decreases significantly as $N_A$ increases because more training data improves the generalization ability of the model. However, as $N_A$ further increases, averaged ratio error increases marginally \revise{and then stabilizes}. This counter-intuitive phenomenon happens often when the distribution of training set is different from the test set. When the training set becomes extremely large, the model learns the very fine-grained details of the training distribution, which tend to not generalize better to the test set.
 
\noindent
\revise{\textbf{Choosing training data size.} Note that the best choice of training data sizes for different columns can be different. Our goal is to train a workload-agnostic estimator, and it can be too expensive (if not impossible) to tune the training data size for each different workload.
From Figure~\ref{fig:err_vs_nd}, which reports the average error across a number of tables for varying $N_A$, a general trend is that: after $N_A$ exceeds some threshold (e.g., $10^4$), the performance of the learned estimator stabilizes. Thus, in practice, we simply choose a training data size that is large enough (exceeding the threshold) but does not make the training data preparation and model training too expensive. We set the training data size $N_A=0.72 \times 10^6$ to train the estimator (available in \cite{url:est}) used in all the other experiments.
%
}

\noindent\revise{
\noindent\textbf{Out-of-range generalization.}
In our training set, the max population size is $10^9$ and the max population NDV is $10^9$. To evaluate the out-of-range performance, \ie, {\em estimation for a column with large NDV not seen at training time}, of our learned estimator, we create such extended test datasets:
%
%
for each column in the original test datasets, we duplicate it $10^4$ times and add a suffix $i$ to each value in the $i$th duplicate. 
In this way, columns in the extended datasets have extremely large population sizes (maximum at $3\times 10^{12}$) and NDVs (maximum at $0.9\times 10^{11}$). 
%
%
We focus on columns with population size in $[10^{10},3\times10^{12}]$ and population NDV in $[10^{10},0.9\times 10^{11}]$, and report the average ratio error for all methods with sampling rate $10^{-3}$ on these columns in \ctab\ref{tbl:out_of_range}. 
Note that both the original and extended test datasets are {\em not} used for training our estimator.
It can be seen that our method also works very well even when NDV to be estimated is out-of-range, and is still better than other baselines. The out-of-range generalizability  of our learned estimator is partly due to the logarithm operations as discussed in \csec\ref{sssec:model_struc}.
%
}

\begin{table}[t]
\vspace{-0mm}
\setlength{\tabcolsep}{0.1em}
\renewcommand{\arraystretch}{0.8}
\caption{\revise{Error on columns with out-of-range NDV ($>10^{10}$)}}
\scalebox{0.98}{
\begin{tabular}{|l|l|l|l|l|l|l|l|l|l|l|}
\hline
           & GEE  & \makecell{HYB\\GEE} & \makecell{HYB\\SKEW} & AE           & Chao         & \makecell{Chao\\-Lee} & \makecell{Shlo\\sser} & APML & Our           \\ \hline
Avg ratio error    & 20.6  & 30.1  & 78.3   & 15.6          & 143.2          & 164.5     & 15.7     &    37.5        & \textbf{4.2}  \\ \hline
\end{tabular}
}
\label{tbl:out_of_range}
\end{table}

%% file: paper-appendix.tex
\section{Appendix}
\subsection{From sample to sample profile}
\label{ssec:sample_vs_profile}
We first formally show that the NDV that maximizes the likelihood of observing the sample $\sample$ is identical to the NDV that maximizes the likelihood of observing the sample profile $f$. Consider two samples $\sample$ and $\sample'$ with the same sample profile $f$, there must exist a ``relabeling'' of values, \ie, a permutation $\pi: \Omega \rightarrow \Omega$ where $\Omega$ is the domain of all possible values, such that $\pi(\sample) = \sample'$. For every possible column $\col$ of $\sample$, we can apply the same relabeling to $\col$. Since relabeling doesn't change the probability of generating the sample, we have 
\begin{equation}
\pr{\sample|\col}=\pr{\pi(\sample)|\pi(\col)}=\pr{\sample'|\pi(\col)}
\end{equation}
For estimators to be applied to any unseen columns with arbitrary values, it is natural to regard that the specific values provide no information on NDV. In other words, relabeling does not change the probability of a column when conditioned on NDV: $\pr{\col|D}=\pr{\pi(\col)|D}$.
Therefore, 
\begin{equation}
\begin{split}
&\pr{\sample|D} = \sum_{\col}\pr{\sample|\col}\pr{\col|D} \\
=& \sum_{\col}\pr{\sample'|\pi(\col)}\pr{\col|D}= \sum_{\col}\pr{\sample'|\pi(\col)}\pr{\pi(\col)|D}\\
=& \sum_{\pi(\col)}\pr{\sample'|\pi(\col)}\pr{\pi(\col)|D} =\pr{\sample'|D}
\end{split}
\end{equation}
This means:
\begin{equation}
\pr{f|D} = \sum_{\sample \in \sample_f} \pr{\sample|D} = |\sample_f|\pr{\sample|D}
\end{equation}
where $\sample_f$ denotes the set of samples whose profile is $f$. Therefore:
\begin{equation}
\arg \max_D \pr{f|D} = \arg \max_D |\sample_f|\pr{\sample|D} = \arg \max_D \pr{\sample|D}
\end{equation}
This means using sample profile is theoretically identical to using sample.
However, using profile is practically advantageous than using sample because profile is a more compact representation and the input dimensionality is reduced. This makes the likelihood function much simpler. In addition, we are to train a model as an estimator and smaller input feature dimensionality makes learning easier.

\subsection{Proof of \cprop\ref{prop:hardness}}
\label{ssec:proof_prop_1}

\begin{manualproposition}{1} \label{prop:app_hardness}
For any size-$n$ sample with profile $f$ and NDV $d$, there exist two size-$N$ columns $\col_1$ and $\col_2$ with NDVs $D_1$ and $D_2$, respectively, such that with probability at least $\gamma$ we cannot distinguish whether the observed $f$ is generated from $\col_1$ or from $\col_2$, with
\begin{equation}\label{equ:app_hardinstance}
D_1 = \left\lfloor \frac{N-n}{4n} \left(\ln(\frac{1}{\gamma}) - \frac{2}{e^c}\right) \right\rfloor + d \hbox{~~and~~}
D_2 = d
\end{equation}
for $\gamma \geq e^{-4n-2e^{-c}}$ and $n \geq d(\ln d + c)$.
\end{manualproposition}

\begin{proof}
Consider two columns (multi-sets) $\col_1 = \col_0 \cup \Delta_1$ and $\col_2 = \col_0 \cup \Delta_2$ sharing the common subset $\col_0$. Suppose $\col_0$ is a multi-set with size $N-k$ and NDV $d$. Let $\Delta_1$ be a multi-set containing $k$ values that are not in $\col_0$, each appearing exactly once in $\Delta_1$, and $\Delta_2$ be a size-$k$ multi-set containing only values that already have appeared in $\col_0$. Thus, the two multi-sets $\col_1 = \col_0 \cup \Delta_1$ and $\col_2 = \col_0 \cup \Delta_2$ have NDVs $D_1 = k + d$ and $D_2 = d$, respectively.

We draw a size-$n$ sample $\sample_1 = \{a_1, \ldots, a_n\}$ from $\col_1$ uniformly at random without replacement\footnote{Analysis for the case with replacement is similar and simpler, and is omitted here.}. When drawing the $i$th value $a_i$, 
\[
\pr{a_i \in \col_0 \mid a_1 \in \col_0, \ldots, a_{i-1} \in \col_0} = \frac{N-k-i+1}{N-i+1},
\]
and thus, the probability that $\sample_1$ contain values only from $\col_0$ is
\begin{small}
\begin{align}
\pr{\sample_1 \subset \col_0} = & \prod_{i=1}^n \pr{a_i \in \col_0 \mid a_1 \in \col_0,\ldots, a_{i-1} \in \col_0} \label{equ:app_hardinstance:sample}
\\
= & \prod_{i=1}^n \frac{N-k-i+1}{N-i+1}\geq (\frac{N-n-k}{N-n})^n = (1-\frac{k}{N-n})^n \nonumber
\end{align}
\end{small}
Similarly, for a size-$n$ sample $\sample_2$ drawn from $\col_2$ uniformly at random, we have $\pr{\sample_2 \subset \col_0} \geq (1-\frac{k}{N-n})^n$.

After a sample profile $f$ is observed, since conditioned on $\sample_1 \subset \col_0$ and $\sample_2 \subset \col_0$, $f_1$ and $f_2$ (profiles of $\sample_1$ and $\sample_2$, respectively) follow the same distribution, \ie, $\pr{f_1 = f \mid \hbox{$\sample_1 \subset \col_0$ and $\sample_2 \subset \col_0$}} = \pr{f_2 = f \mid \hbox{$\sample_1 \subset \col_0$ and $\sample_2 \subset \col_0$}}$, we cannot distinguish whether the underlying data column is $\col_1$ or $\col_2$. From \eqref{equ:app_hardinstance:sample}, we have
\[
\pr{\hbox{$\sample_1 \subset \col_0$ and $\sample_2 \subset \col_0$}} \geq (1-\frac{k}{N-n})^{2n}\geq e^{\frac{-4kn}{N-n}}
\]
where the last inequality follows from the claim that $1-z \geq e^{-2z}$ for $z = k/(N-n) \leq 1/2$, which can be easily verified.

%
Moreover, we want to ensure that $\sample_1$ and $\sample_2$ both cover all the distinct values of $\col_0$, \ie, sample NDVs and NDV of $\col_0$ are all $d$. Suppose each distinct value appears with the same frequency in $\col_0$. From the Coupon Collector's problem (\csec~3.6.2 in \cite{motwanir1995ra}), conditioned on $\sample_1 \subset \col_0$, $\sample_2 \subset \col_0$, and $n \geq d(\ln d + c)$, with probability at least $e^{-e^{-c}}$ (for $c>0$), a size-$n$ sample $\sample_1$ (or $\sample_2$) contains each of the $d$ distinct values in $\col_0$ at least once. Therefore, we have
\[
\pr{\hbox{$\sample_i \subset \col_0$ and sample NDV is $d$ in $\sample_i$ for $i=1,2$}} \geq e^{\frac{-4kn}{N-n}} \cdot e^{-2e^{-c}}.
\]

Let $\gamma = e^{\frac{-4kn}{N-n}} \cdot e^{-2e^{-c}}$ and thus $k =\frac{N-n}{4n} \left(\ln(\frac{1}{\gamma}) - \frac{2}{e^c}\right)$.


For $\gamma \geq e^{-4n} \cdot e^{-2e^{-c}}$ (which ensures $|\col_0| = N-k \geq n$, \ie, a sufficient number of values can be drawn from $\col_0$ into $\sample_1$ or $\sample_2$) and $z = k/(N-n) \leq 1/2$ (from our choice of $k$ and $\gamma$), we can conclude that, with probability at least $\gamma$, we cannot distinguish whether $f$ is from $\col_1$ or from $\col_2$, with desired NDVs $D_1$ and $D_2$ in \eqref{equ:app_hardinstance}.
\end{proof}

\subsection{Proof of Theorem~\ref{theorem:b2}}
\label{ssec:proof_b2}
\begin{manualtheorem}{1}
For any size-$n$ sample with profile $f$ and sample NDV $d$ observed (where $n \geq d(\ln d + c)$), with probability $\gamma \geq e^{-4n-2e^{-c}}$, there exist a choice of column with NDV $D$ such that any estimation of $D$ based on $f$, \ie, $\est(f, N)$, has ratio error at least
\begin{equation}\label{eq:app_b2}
\err(\est(f, N) ,D) \geq \sqrt{\frac{\left\lfloor \frac{N-n}{4n} \left(\ln(\frac{1}{\gamma}) - \frac{2}{e^c}\right) \right\rfloor + d}{d}} \triangleq b(d,n).
\end{equation}
\end{manualtheorem}
\begin{proof}
Consider the two choices of columns $\col_1$ and $\col_2$ constructed in \cprop\ref{prop:app_hardness}, with NDVs $D_1$ and $D_2$, respectively, as in \eqref{equ:app_hardinstance}.
Since with probability at least $\gamma$, we cannot distinguish whether $f$ is from $\col_1$ or $\col_2$, for ${\hat D} = \est(f,N)$, the ratio error is at least
$\max(\err({\hat D}, D_1), \err({\hat D}, D_2))$, which is minimized when ${\hat D} = \sqrt{D_1 D_2}$. And thus, the lower bound $b(d,n)$ of ratio error in \eqref{eq:app_b2} follows for the choice of $D_1$ and $D_2$ in \eqref{equ:app_hardinstance}.
%
\end{proof}

%% file: main.bbl

\begin{thebibliography}{60}


\ifx \showCODEN    \undefined \def \showCODEN     #1{\unskip}     \fi
\ifx \showDOI      \undefined \def \showDOI       #1{#1}\fi
\ifx \showISBNx    \undefined \def \showISBNx     #1{\unskip}     \fi
\ifx \showISBNxiii \undefined \def \showISBNxiii  #1{\unskip}     \fi
\ifx \showISSN     \undefined \def \showISSN      #1{\unskip}     \fi
\ifx \showLCCN     \undefined \def \showLCCN      #1{\unskip}     \fi
\ifx \shownote     \undefined \def \shownote      #1{#1}          \fi
\ifx \showarticletitle \undefined \def \showarticletitle #1{#1}   \fi
\ifx \showURL      \undefined \def \showURL       {\relax}        \fi
\providecommand\bibfield[2]{#2}
\providecommand\bibinfo[2]{#2}
\providecommand\natexlab[1]{#1}
\providecommand\showeprint[2][]{arXiv:#2}

\bibitem[\protect\citeauthoryear{??}{air}{2020}]%
        {airline_data}
 \bibinfo{year}{2020}\natexlab{}.
\newblock \bibinfo{title}{Airlines Departure Delay}.
\newblock
\newblock
\urldef\tempurl%
\url{https://www.openml.org/d/42728}
\showURL{%
\tempurl}


\bibitem[\protect\citeauthoryear{??}{box}{2020}]%
        {box_plot}
 \bibinfo{year}{2020}\natexlab{}.
\newblock \bibinfo{title}{Box plot}.
\newblock
\newblock
\urldef\tempurl%
\url{https://en.wikipedia.org/wiki/Box_plot}
\showURL{%
\tempurl}


\bibitem[\protect\citeauthoryear{??}{bts}{2020}]%
        {bts}
 \bibinfo{year}{2020}\natexlab{}.
\newblock \bibinfo{title}{Bureau of Transportation Statistics}.
\newblock
\newblock
\urldef\tempurl%
\url{https://www.transtats.bts.gov/}
\showURL{%
\tempurl}


\bibitem[\protect\citeauthoryear{??}{Cam}{2020}]%
        {Campaign_finance}
 \bibinfo{year}{2020}\natexlab{}.
\newblock \bibinfo{title}{Campaign finance data}.
\newblock
\newblock
\urldef\tempurl%
\url{https://www.fec.gov/data/}
\showURL{%
\tempurl}


\bibitem[\protect\citeauthoryear{??}{dat}{2020}]%
        {data_dmv}
 \bibinfo{year}{2020}\natexlab{}.
\newblock \bibinfo{title}{Department of Motor Vehicle (DMV) Office Locations}.
\newblock
\newblock
\urldef\tempurl%
\url{https://catalog.data.gov/dataset/department-of-motor-vehicle-dmv-office-locations}
\showURL{%
\tempurl}


\bibitem[\protect\citeauthoryear{??}{lea}{2020}]%
        {leakyrelu}
 \bibinfo{year}{2020}\natexlab{}.
\newblock \bibinfo{title}{Leaky ReLU}.
\newblock
\newblock
\urldef\tempurl%
\url{https://pytorch.org/docs/stable/generated/torch.nn.LeakyReLU.html}
\showURL{%
\tempurl}


\bibitem[\protect\citeauthoryear{??}{max}{2020}]%
        {maxcompute}
 \bibinfo{year}{2020}\natexlab{}.
\newblock \bibinfo{title}{MaxCompute}.
\newblock
\newblock
\urldef\tempurl%
\url{https://www.alibabacloud.com/product/maxcompute}
\showURL{%
\tempurl}


\bibitem[\protect\citeauthoryear{??}{pyd}{2020}]%
        {pydistinct}
 \bibinfo{year}{2020}\natexlab{}.
\newblock \bibinfo{title}{Pydistinct - Population Distinct Value Estimators}.
\newblock
\newblock
\urldef\tempurl%
\url{https://pydistinct.readthedocs.io/}
\showURL{%
\tempurl}


\bibitem[\protect\citeauthoryear{??}{ran}{2020}]%
        {random_fixed_sum}
 \bibinfo{year}{2020}\natexlab{}.
\newblock \bibinfo{title}{Random numbers that add to 100: Matlab}.
\newblock
\newblock
\urldef\tempurl%
\url{https://stackoverflow.com/questions/8064629/random-numbers-that-add-to-100-matlab}
\showURL{%
\tempurl}


\bibitem[\protect\citeauthoryear{??}{bre}{2020}]%
        {brentq}
 \bibinfo{year}{2020}\natexlab{}.
\newblock \bibinfo{title}{scipy.optimize.brentq}.
\newblock
\newblock
\urldef\tempurl%
\url{https://docs.scipy.org/doc/scipy/reference/generated/scipy.optimize.brentq.html}
\showURL{%
\tempurl}


\bibitem[\protect\citeauthoryear{??}{sko}{2020}]%
        {skorch}
 \bibinfo{year}{2020}\natexlab{}.
\newblock \bibinfo{title}{skorch documentation}.
\newblock
\newblock
\urldef\tempurl%
\url{https://skorch.readthedocs.io/en/stable/}
\showURL{%
\tempurl}


\bibitem[\protect\citeauthoryear{??}{nvc}{2020}]%
        {nvcr}
 \bibinfo{year}{2020}\natexlab{}.
\newblock \bibinfo{title}{Voter Registration Statistics}.
\newblock
\newblock
\urldef\tempurl%
\url{https://www.ncsbe.gov/results-data/voter-registration-data}
\showURL{%
\tempurl}


\bibitem[\protect\citeauthoryear{??}{met}{2020}]%
        {metacademy_wd}
 \bibinfo{year}{2020}\natexlab{}.
\newblock \bibinfo{title}{weight decay in neural networks}.
\newblock
\newblock
\urldef\tempurl%
\url{https://metacademy.org/graphs/concepts/weight_decay_neural_networks}
\showURL{%
\tempurl}


\bibitem[\protect\citeauthoryear{??}{rfs}{2021}]%
        {rfs_matlab}
 \bibinfo{year}{2021}\natexlab{}.
\newblock \bibinfo{title}{{Random Vectors with Fixed Sum - File Exchange -
  MATLAB Central}}.
\newblock
\newblock
\urldef\tempurl%
\url{https://www.mathworks.com/matlabcentral/fileexchange/9700-random-vectors-with-fixed-sum}
\showURL{%
\tempurl}
\newblock
\shownote{[Online; accessed 27. Apr. 2021].}


\bibitem[\protect\citeauthoryear{Anagnostopoulos and
  Triantafillou}{Anagnostopoulos and Triantafillou}{2015}]%
        {anagnostopoulos2015learning}
\bibfield{author}{\bibinfo{person}{Christos Anagnostopoulos} {and}
  \bibinfo{person}{Peter Triantafillou}.} \bibinfo{year}{2015}\natexlab{}.
\newblock \showarticletitle{Learning to accurately count with query-driven
  predictive analytics}. In \bibinfo{booktitle}{\emph{2015 IEEE international
  conference on big data (big data)}}. IEEE, \bibinfo{pages}{14--23}.
\newblock


\bibitem[\protect\citeauthoryear{Bishop}{Bishop}{2006}]%
        {bishop2006pattern}
\bibfield{author}{\bibinfo{person}{Christopher~M Bishop}.}
  \bibinfo{year}{2006}\natexlab{}.
\newblock \bibinfo{booktitle}{\emph{Pattern recognition and machine learning}}.
\newblock \bibinfo{publisher}{springer}.
\newblock


\bibitem[\protect\citeauthoryear{Brent}{Brent}{1973}]%
        {brent1973algorithms}
\bibfield{author}{\bibinfo{person}{Richard~P Brent}.}
  \bibinfo{year}{1973}\natexlab{}.
\newblock \bibinfo{title}{Algorithms for Minimization without Derivatives,
  chap. 4}.
\newblock
\newblock


\bibitem[\protect\citeauthoryear{Bunge and Fitzpatrick}{Bunge and
  Fitzpatrick}{1993}]%
        {bunge1993estimating}
\bibfield{author}{\bibinfo{person}{John Bunge} {and} \bibinfo{person}{Michael
  Fitzpatrick}.} \bibinfo{year}{1993}\natexlab{}.
\newblock \showarticletitle{Estimating the number of species: a review}.
\newblock \bibinfo{journal}{\emph{J. Amer. Statist. Assoc.}}
  \bibinfo{volume}{88}, \bibinfo{number}{421} (\bibinfo{year}{1993}),
  \bibinfo{pages}{364--373}.
\newblock


\bibitem[\protect\citeauthoryear{Chambers, Steel, Wang, and Welsh}{Chambers
  et~al\mbox{.}}{2012}]%
        {chambers2012maximum}
\bibfield{author}{\bibinfo{person}{Raymond~L Chambers},
  \bibinfo{person}{David~G Steel}, \bibinfo{person}{Suojin Wang}, {and}
  \bibinfo{person}{Alan Welsh}.} \bibinfo{year}{2012}\natexlab{}.
\newblock \bibinfo{booktitle}{\emph{Maximum likelihood estimation for sample
  surveys}}.
\newblock \bibinfo{publisher}{CRC Press}.
\newblock


\bibitem[\protect\citeauthoryear{Chao}{Chao}{1984}]%
        {chao1984nonparametric}
\bibfield{author}{\bibinfo{person}{Anne Chao}.}
  \bibinfo{year}{1984}\natexlab{}.
\newblock \showarticletitle{Nonparametric estimation of the number of classes
  in a population}.
\newblock \bibinfo{journal}{\emph{Scandinavian Journal of statistics}}
  (\bibinfo{year}{1984}), \bibinfo{pages}{265--270}.
\newblock


\bibitem[\protect\citeauthoryear{Chao and Lee}{Chao and Lee}{1992}]%
        {chao1992estimating}
\bibfield{author}{\bibinfo{person}{Anne Chao} {and} \bibinfo{person}{Shen-Ming
  Lee}.} \bibinfo{year}{1992}\natexlab{}.
\newblock \showarticletitle{Estimating the number of classes via sample
  coverage}.
\newblock \bibinfo{journal}{\emph{Journal of the American statistical
  Association}} \bibinfo{volume}{87}, \bibinfo{number}{417}
  (\bibinfo{year}{1992}), \bibinfo{pages}{210--217}.
\newblock


\bibitem[\protect\citeauthoryear{Charikar, Chaudhuri, Motwani, and
  Narasayya}{Charikar et~al\mbox{.}}{2000}]%
        {charikar2000towards}
\bibfield{author}{\bibinfo{person}{Moses Charikar}, \bibinfo{person}{Surajit
  Chaudhuri}, \bibinfo{person}{Rajeev Motwani}, {and} \bibinfo{person}{Vivek
  Narasayya}.} \bibinfo{year}{2000}\natexlab{}.
\newblock \showarticletitle{Towards estimation error guarantees for distinct
  values}. In \bibinfo{booktitle}{\emph{Proceedings of the nineteenth ACM
  SIGMOD-SIGACT-SIGART symposium on Principles of database systems}}.
  \bibinfo{pages}{268--279}.
\newblock


\bibitem[\protect\citeauthoryear{Charikar, Shiragur, and Sidford}{Charikar
  et~al\mbox{.}}{2019}]%
        {charikar2019efficient}
\bibfield{author}{\bibinfo{person}{Moses Charikar}, \bibinfo{person}{Kirankumar
  Shiragur}, {and} \bibinfo{person}{Aaron Sidford}.}
  \bibinfo{year}{2019}\natexlab{}.
\newblock \showarticletitle{Efficient profile maximum likelihood for universal
  symmetric property estimation}. In \bibinfo{booktitle}{\emph{Proceedings of
  the 51st Annual ACM SIGACT Symposium on Theory of Computing}}.
  \bibinfo{pages}{780--791}.
\newblock


\bibitem[\protect\citeauthoryear{Chaudhuri, Ding, and Kandula}{Chaudhuri
  et~al\mbox{.}}{2017}]%
        {sigmod:ChaudhuriDK17}
\bibfield{author}{\bibinfo{person}{Surajit Chaudhuri}, \bibinfo{person}{Bolin
  Ding}, {and} \bibinfo{person}{Srikanth Kandula}.}
  \bibinfo{year}{2017}\natexlab{}.
\newblock \showarticletitle{Approximate Query Processing: No Silver Bullet}. In
  \bibinfo{booktitle}{\emph{{SIGMOD}}}. \bibinfo{pages}{511--519}.
\newblock


\bibitem[\protect\citeauthoryear{Cohen and Nezri}{Cohen and Nezri}{2019}]%
        {cohen2019cardinality}
\bibfield{author}{\bibinfo{person}{Reuven Cohen} {and} \bibinfo{person}{Yuval
  Nezri}.} \bibinfo{year}{2019}\natexlab{}.
\newblock \showarticletitle{Cardinality Estimation in a Virtualized Network
  Device Using Online Machine Learning}.
\newblock \bibinfo{journal}{\emph{IEEE/ACM Transactions on Networking}}
  \bibinfo{volume}{27}, \bibinfo{number}{5} (\bibinfo{year}{2019}),
  \bibinfo{pages}{2098--2110}.
\newblock


\bibitem[\protect\citeauthoryear{Dutt, Wang, Narasayya, and Chaudhuri}{Dutt
  et~al\mbox{.}}{2020}]%
        {pvldb:DuttWNC20}
\bibfield{author}{\bibinfo{person}{Anshuman Dutt}, \bibinfo{person}{Chi Wang},
  \bibinfo{person}{Vivek~R. Narasayya}, {and} \bibinfo{person}{Surajit
  Chaudhuri}.} \bibinfo{year}{2020}\natexlab{}.
\newblock \showarticletitle{Efficiently Approximating Selectivity Functions
  using Low Overhead Regression Models}.
\newblock \bibinfo{journal}{\emph{Proc. {VLDB} Endow.}} \bibinfo{volume}{13},
  \bibinfo{number}{11} (\bibinfo{year}{2020}), \bibinfo{pages}{2215--2228}.
\newblock


\bibitem[\protect\citeauthoryear{Dutt, Wang, Nazi, Kandula, Narasayya, and
  Chaudhuri}{Dutt et~al\mbox{.}}{2019}]%
        {dutt2019selectivity}
\bibfield{author}{\bibinfo{person}{Anshuman Dutt}, \bibinfo{person}{Chi Wang},
  \bibinfo{person}{Azade Nazi}, \bibinfo{person}{Srikanth Kandula},
  \bibinfo{person}{Vivek Narasayya}, {and} \bibinfo{person}{Surajit
  Chaudhuri}.} \bibinfo{year}{2019}\natexlab{}.
\newblock \showarticletitle{Selectivity estimation for range predicates using
  lightweight models}.
\newblock \bibinfo{journal}{\emph{Proc. {VLDB} Endow.}} \bibinfo{volume}{12},
  \bibinfo{number}{9} (\bibinfo{year}{2019}), \bibinfo{pages}{1044--1057}.
\newblock


\bibitem[\protect\citeauthoryear{Flajolet, Fusy, Gandouet, and
  Meunier}{Flajolet et~al\mbox{.}}{2007}]%
        {flajolet2007hyperloglog}
\bibfield{author}{\bibinfo{person}{Philippe Flajolet},
  \bibinfo{person}{{\'E}ric Fusy}, \bibinfo{person}{Olivier Gandouet}, {and}
  \bibinfo{person}{Fr{\'e}d{\'e}ric Meunier}.} \bibinfo{year}{2007}\natexlab{}.
\newblock \showarticletitle{Hyperloglog: the analysis of a near-optimal
  cardinality estimation algorithm}. In \bibinfo{booktitle}{\emph{Proceedings
  of the Analysis of Algorithms Conference}}. \bibinfo{pages}{137--156}.
\newblock


\bibitem[\protect\citeauthoryear{Goodfellow, Bengio, and Courville}{Goodfellow
  et~al\mbox{.}}{2017}]%
        {goodfellow_bengio_courville_2017}
\bibfield{author}{\bibinfo{person}{Ian Goodfellow}, \bibinfo{person}{Yoshua
  Bengio}, {and} \bibinfo{person}{Aaron Courville}.}
  \bibinfo{year}{2017}\natexlab{}.
\newblock \bibinfo{booktitle}{\emph{Deep learning Ch. 5 Machine Learning
  Basics}}.
\newblock \bibinfo{publisher}{The MIT Press}, \bibinfo{pages}{132–133}.
\newblock


\bibitem[\protect\citeauthoryear{Haas, Naughton, Seshadri, and Stokes}{Haas
  et~al\mbox{.}}{1995}]%
        {haas1995sampling}
\bibfield{author}{\bibinfo{person}{Peter~J Haas}, \bibinfo{person}{Jeffrey~F
  Naughton}, \bibinfo{person}{S Seshadri}, {and} \bibinfo{person}{Lynne
  Stokes}.} \bibinfo{year}{1995}\natexlab{}.
\newblock \showarticletitle{Sampling-based estimation of the number of distinct
  values of an attribute}. In \bibinfo{booktitle}{\emph{VLDB}},
  Vol.~\bibinfo{volume}{95}. \bibinfo{pages}{311--322}.
\newblock


\bibitem[\protect\citeauthoryear{Haas and Stokes}{Haas and Stokes}{1998}]%
        {haas1998estimating}
\bibfield{author}{\bibinfo{person}{Peter~J Haas} {and} \bibinfo{person}{Lynne
  Stokes}.} \bibinfo{year}{1998}\natexlab{}.
\newblock \showarticletitle{Estimating the number of classes in a finite
  population}.
\newblock \bibinfo{journal}{\emph{J. Amer. Statist. Assoc.}}
  \bibinfo{volume}{93}, \bibinfo{number}{444} (\bibinfo{year}{1998}),
  \bibinfo{pages}{1475--1487}.
\newblock


\bibitem[\protect\citeauthoryear{Hao and Orlitsky}{Hao and Orlitsky}{2019}]%
        {hao2019broad}
\bibfield{author}{\bibinfo{person}{Yi Hao} {and} \bibinfo{person}{Alon
  Orlitsky}.} \bibinfo{year}{2019}\natexlab{}.
\newblock \showarticletitle{The broad optimality of profile maximum
  likelihood}. In \bibinfo{booktitle}{\emph{Advances in Neural Information
  Processing Systems}}. \bibinfo{pages}{10991--11003}.
\newblock


\bibitem[\protect\citeauthoryear{Harmouch and Naumann}{Harmouch and
  Naumann}{2017}]%
        {harmouch2017cardinality}
\bibfield{author}{\bibinfo{person}{Hazar Harmouch} {and} \bibinfo{person}{Felix
  Naumann}.} \bibinfo{year}{2017}\natexlab{}.
\newblock \showarticletitle{Cardinality estimation: An experimental survey}.
\newblock \bibinfo{journal}{\emph{Proceedings of the VLDB Endowment}}
  \bibinfo{volume}{11}, \bibinfo{number}{4} (\bibinfo{year}{2017}),
  \bibinfo{pages}{499--512}.
\newblock


\bibitem[\protect\citeauthoryear{Haykin}{Haykin}{1998}]%
        {haykin1998nn}
\bibfield{author}{\bibinfo{person}{Simon Haykin}.}
  \bibinfo{year}{1998}\natexlab{}.
\newblock \bibinfo{booktitle}{\emph{Neural Networks: A Comprehensive
  Foundation}}.
\newblock \bibinfo{publisher}{Prentice Hall}.
\newblock


\bibitem[\protect\citeauthoryear{Hilprecht, Schmidt, Kulessa, Molina, Kersting,
  and Binnig}{Hilprecht et~al\mbox{.}}{2020}]%
        {hilprecht2020deepdb}
\bibfield{author}{\bibinfo{person}{Benjamin Hilprecht},
  \bibinfo{person}{Andreas Schmidt}, \bibinfo{person}{Moritz Kulessa},
  \bibinfo{person}{Alejandro Molina}, \bibinfo{person}{Kristian Kersting},
  {and} \bibinfo{person}{Carsten Binnig}.} \bibinfo{year}{2020}\natexlab{}.
\newblock \showarticletitle{DeepDB: learn from data, not from queries!}
\newblock \bibinfo{journal}{\emph{Proceedings of the VLDB Endowment}}
  \bibinfo{volume}{13}, \bibinfo{number}{7} (\bibinfo{year}{2020}),
  \bibinfo{pages}{992--1005}.
\newblock


\bibitem[\protect\citeauthoryear{Hines}{Hines}{1996}]%
        {hines1996logarithmic}
\bibfield{author}{\bibinfo{person}{J~Wesley Hines}.}
  \bibinfo{year}{1996}\natexlab{}.
\newblock \showarticletitle{A logarithmic neural network architecture for
  unbounded non-linear function approximation}. In
  \bibinfo{booktitle}{\emph{Proceedings of International Conference on Neural
  Networks (ICNN'96)}}, Vol.~\bibinfo{volume}{2}. IEEE,
  \bibinfo{pages}{1245--1250}.
\newblock


\bibitem[\protect\citeauthoryear{Ioffe and Szegedy}{Ioffe and Szegedy}{2015}]%
        {ioffe2015batch}
\bibfield{author}{\bibinfo{person}{Sergey Ioffe} {and}
  \bibinfo{person}{Christian Szegedy}.} \bibinfo{year}{2015}\natexlab{}.
\newblock \showarticletitle{Batch Normalization: Accelerating Deep Network
  Training by Reducing Internal Covariate Shift}. In
  \bibinfo{booktitle}{\emph{Proceedings of the 32nd International Conference on
  International Conference on Machine Learning - Volume 37}} (Lille, France)
  \emph{(\bibinfo{series}{ICML'15})}. \bibinfo{publisher}{JMLR.org},
  \bibinfo{pages}{448–456}.
\newblock


\bibitem[\protect\citeauthoryear{Juszczak, Tax, and Duin}{Juszczak
  et~al\mbox{.}}{[n.d.]}]%
        {juszczak2002feature}
\bibfield{author}{\bibinfo{person}{Piotr Juszczak}, \bibinfo{person}{D Tax},
  {and} \bibinfo{person}{Robert~PW Duin}.} \bibinfo{year}{[n.d.]}\natexlab{}.
\newblock \showarticletitle{Feature scaling in support vector data
  description}. Citeseer.
\newblock


\bibitem[\protect\citeauthoryear{Kiefer, Heimel, Bre{\ss}, and Markl}{Kiefer
  et~al\mbox{.}}{2017}]%
        {kiefer2017estimating}
\bibfield{author}{\bibinfo{person}{Martin Kiefer}, \bibinfo{person}{Max
  Heimel}, \bibinfo{person}{Sebastian Bre{\ss}}, {and} \bibinfo{person}{Volker
  Markl}.} \bibinfo{year}{2017}\natexlab{}.
\newblock \showarticletitle{Estimating join selectivities using
  bandwidth-optimized kernel density models}.
\newblock \bibinfo{journal}{\emph{Proceedings of the VLDB Endowment}}
  \bibinfo{volume}{10}, \bibinfo{number}{13} (\bibinfo{year}{2017}),
  \bibinfo{pages}{2085--2096}.
\newblock


\bibitem[\protect\citeauthoryear{Kingma and Ba}{Kingma and Ba}{2015}]%
        {kingma2014adam}
\bibfield{author}{\bibinfo{person}{Diederik~P. Kingma} {and}
  \bibinfo{person}{Jimmy Ba}.} \bibinfo{year}{2015}\natexlab{}.
\newblock \showarticletitle{Adam: {A} Method for Stochastic Optimization}. In
  \bibinfo{booktitle}{\emph{3rd International Conference on Learning
  Representations, {ICLR} 2015, San Diego, CA, USA, May 7-9, 2015, Conference
  Track Proceedings}}, \bibfield{editor}{\bibinfo{person}{Yoshua Bengio} {and}
  \bibinfo{person}{Yann LeCun}} (Eds.).
\newblock
\urldef\tempurl%
\url{http://arxiv.org/abs/1412.6980}
\showURL{%
\tempurl}


\bibitem[\protect\citeauthoryear{Kipf, Kipf, Radke, Leis, Boncz, and
  Kemper}{Kipf et~al\mbox{.}}{2018}]%
        {kipf2018learned}
\bibfield{author}{\bibinfo{person}{Andreas Kipf}, \bibinfo{person}{Thomas
  Kipf}, \bibinfo{person}{Bernhard Radke}, \bibinfo{person}{Viktor Leis},
  \bibinfo{person}{Peter Boncz}, {and} \bibinfo{person}{Alfons Kemper}.}
  \bibinfo{year}{2018}\natexlab{}.
\newblock \showarticletitle{Learned cardinalities: Estimating correlated joins
  with deep learning}.
\newblock \bibinfo{journal}{\emph{arXiv preprint arXiv:1809.00677}}
  (\bibinfo{year}{2018}).
\newblock


\bibitem[\protect\citeauthoryear{Lakshmi and Zhou}{Lakshmi and Zhou}{1998}]%
        {lakshmi1998selectivity}
\bibfield{author}{\bibinfo{person}{Seetha Lakshmi} {and}
  \bibinfo{person}{Shaoyu Zhou}.} \bibinfo{year}{1998}\natexlab{}.
\newblock \showarticletitle{Selectivity estimation in extensible databases-a
  neural network approach}. In \bibinfo{booktitle}{\emph{VLDB}},
  Vol.~\bibinfo{volume}{98}. \bibinfo{pages}{24--27}.
\newblock


\bibitem[\protect\citeauthoryear{Library}{Library}{2021}]%
        {url:est}
\bibfield{author}{\bibinfo{person}{Library}.} \bibinfo{year}{2021}\natexlab{}.
\newblock \bibinfo{title}{{An learned sample-based NDV estimator}}.
\newblock
  \bibinfo{howpublished}{\url{https://github.com/wurenzhi/learned_ndv_estimator}}.
\newblock
\newblock
\shownote{[Online; accessed 11-October-2021].}


\bibitem[\protect\citeauthoryear{Liu, Xu, Yu, Corvinelli, and Zuzarte}{Liu
  et~al\mbox{.}}{2015}]%
        {liu2015cardinality}
\bibfield{author}{\bibinfo{person}{Henry Liu}, \bibinfo{person}{Mingbin Xu},
  \bibinfo{person}{Ziting Yu}, \bibinfo{person}{Vincent Corvinelli}, {and}
  \bibinfo{person}{Calisto Zuzarte}.} \bibinfo{year}{2015}\natexlab{}.
\newblock \showarticletitle{Cardinality estimation using neural networks}. In
  \bibinfo{booktitle}{\emph{Proceedings of the 25th Annual International
  Conference on Computer Science and Software Engineering}}.
  \bibinfo{pages}{53--59}.
\newblock


\bibitem[\protect\citeauthoryear{Metwally, Agrawal, and Abbadi}{Metwally
  et~al\mbox{.}}{2008}]%
        {metwally2008go}
\bibfield{author}{\bibinfo{person}{Ahmed Metwally}, \bibinfo{person}{Divyakant
  Agrawal}, {and} \bibinfo{person}{Amr~El Abbadi}.}
  \bibinfo{year}{2008}\natexlab{}.
\newblock \showarticletitle{Why go logarithmic if we can go linear? Towards
  effective distinct counting of search traffic}. In
  \bibinfo{booktitle}{\emph{Proceedings of the 11th international conference on
  Extending database technology: Advances in database technology}}.
  \bibinfo{pages}{618--629}.
\newblock


\bibitem[\protect\citeauthoryear{Mohamadi, Khan, and Birol}{Mohamadi
  et~al\mbox{.}}{2017}]%
        {mohamadi2017ntcard}
\bibfield{author}{\bibinfo{person}{Hamid Mohamadi}, \bibinfo{person}{Hamza
  Khan}, {and} \bibinfo{person}{Inanc Birol}.} \bibinfo{year}{2017}\natexlab{}.
\newblock \showarticletitle{ntCard: a streaming algorithm for cardinality
  estimation in genomics data}.
\newblock \bibinfo{journal}{\emph{Bioinformatics}} \bibinfo{volume}{33},
  \bibinfo{number}{9} (\bibinfo{year}{2017}), \bibinfo{pages}{1324--1330}.
\newblock


\bibitem[\protect\citeauthoryear{Motwani and Vassilvitskii}{Motwani and
  Vassilvitskii}{2006}]%
        {motwani2006distinct}
\bibfield{author}{\bibinfo{person}{Rajeev Motwani} {and}
  \bibinfo{person}{Sergei Vassilvitskii}.} \bibinfo{year}{2006}\natexlab{}.
\newblock \showarticletitle{Distinct values estimators for power law
  distributions}. In \bibinfo{booktitle}{\emph{2006 Proceedings of the Third
  Workshop on Analytic Algorithmics and Combinatorics (ANALCO)}}. SIAM,
  \bibinfo{pages}{230--237}.
\newblock


\bibitem[\protect\citeauthoryear{Nath, Gibbons, Seshan, and Anderson}{Nath
  et~al\mbox{.}}{2008}]%
        {nath2008synopsis}
\bibfield{author}{\bibinfo{person}{Suman Nath}, \bibinfo{person}{Phillip~B
  Gibbons}, \bibinfo{person}{Srinivasan Seshan}, {and} \bibinfo{person}{Zachary
  Anderson}.} \bibinfo{year}{2008}\natexlab{}.
\newblock \showarticletitle{Synopsis diffusion for robust aggregation in sensor
  networks}.
\newblock \bibinfo{journal}{\emph{ACM Transactions on Sensor Networks (TOSN)}}
  \bibinfo{volume}{4}, \bibinfo{number}{2} (\bibinfo{year}{2008}),
  \bibinfo{pages}{1--40}.
\newblock


\bibitem[\protect\citeauthoryear{O’Neil, O’Neil, and Chen}{O’Neil
  et~al\mbox{.}}{2007}]%
        {o2007star}
\bibfield{author}{\bibinfo{person}{Patrick~E O’Neil},
  \bibinfo{person}{Elizabeth~J O’Neil}, {and} \bibinfo{person}{Xuedong
  Chen}.} \bibinfo{year}{2007}\natexlab{}.
\newblock \bibinfo{title}{The star schema benchmark (SSB)}.
\newblock
\newblock


\bibitem[\protect\citeauthoryear{Pavlichin, Jiao, and Weissman}{Pavlichin
  et~al\mbox{.}}{2019}]%
        {pavlichin2019approximate}
\bibfield{author}{\bibinfo{person}{Dmitri~S Pavlichin},
  \bibinfo{person}{Jiantao Jiao}, {and} \bibinfo{person}{Tsachy Weissman}.}
  \bibinfo{year}{2019}\natexlab{}.
\newblock \showarticletitle{Approximate Profile Maximum Likelihood.}
\newblock \bibinfo{journal}{\emph{Journal of Machine Learning Research}}
  \bibinfo{volume}{20}, \bibinfo{number}{122} (\bibinfo{year}{2019}),
  \bibinfo{pages}{1--55}.
\newblock
\urldef\tempurl%
\url{http://jmlr.org/papers/v20/18-075.html}
\showURL{%
\tempurl}


\bibitem[\protect\citeauthoryear{Raghu, Poole, Kleinberg, Ganguli, and
  Sohl{-}Dickstein}{Raghu et~al\mbox{.}}{2017}]%
        {icml:RaghuPKGS17}
\bibfield{author}{\bibinfo{person}{Maithra Raghu}, \bibinfo{person}{Ben Poole},
  \bibinfo{person}{Jon~M. Kleinberg}, \bibinfo{person}{Surya Ganguli}, {and}
  \bibinfo{person}{Jascha Sohl{-}Dickstein}.} \bibinfo{year}{2017}\natexlab{}.
\newblock \showarticletitle{On the Expressive Power of Deep Neural Networks}.
  In \bibinfo{booktitle}{\emph{Proceedings of the 34th International Conference
  on Machine Learning, {ICML} 2017}}. \bibinfo{pages}{2847--2854}.
\newblock


\bibitem[\protect\citeauthoryear{Shlosser}{Shlosser}{1981}]%
        {shlosser1981estimation}
\bibfield{author}{\bibinfo{person}{A Shlosser}.}
  \bibinfo{year}{1981}\natexlab{}.
\newblock \showarticletitle{On estimation of the size of the dictionary of a
  long text on the basis of a sample}.
\newblock \bibinfo{journal}{\emph{Engineering Cybernetics}}
  \bibinfo{volume}{19}, \bibinfo{number}{1} (\bibinfo{year}{1981}),
  \bibinfo{pages}{97--102}.
\newblock


\bibitem[\protect\citeauthoryear{Shorten and Khoshgoftaar}{Shorten and
  Khoshgoftaar}{2019}]%
        {shorten2019survey}
\bibfield{author}{\bibinfo{person}{Connor Shorten} {and}
  \bibinfo{person}{Taghi~M Khoshgoftaar}.} \bibinfo{year}{2019}\natexlab{}.
\newblock \showarticletitle{A survey on image data augmentation for deep
  learning}.
\newblock \bibinfo{journal}{\emph{Journal of Big Data}} \bibinfo{volume}{6},
  \bibinfo{number}{1} (\bibinfo{year}{2019}), \bibinfo{pages}{60}.
\newblock


\bibitem[\protect\citeauthoryear{Sidana, Laclau, Amini, Vandelle, and
  Bois-Crettez}{Sidana et~al\mbox{.}}{2017}]%
        {sidana2017kasandr}
\bibfield{author}{\bibinfo{person}{Sumit Sidana}, \bibinfo{person}{Charlotte
  Laclau}, \bibinfo{person}{Massih~R Amini}, \bibinfo{person}{Gilles Vandelle},
  {and} \bibinfo{person}{Andr{\'e} Bois-Crettez}.}
  \bibinfo{year}{2017}\natexlab{}.
\newblock \showarticletitle{KASANDR: a large-scale dataset with implicit
  feedback for recommendation}. In \bibinfo{booktitle}{\emph{Proceedings of the
  40th International ACM SIGIR Conference on Research and Development in
  Information Retrieval}}. \bibinfo{pages}{1245--1248}.
\newblock


\bibitem[\protect\citeauthoryear{Ting}{Ting}{2019}]%
        {ting2019approximate}
\bibfield{author}{\bibinfo{person}{Daniel Ting}.}
  \bibinfo{year}{2019}\natexlab{}.
\newblock \showarticletitle{Approximate Distinct Counts for Billions of
  Datasets}. In \bibinfo{booktitle}{\emph{Proceedings of the 2019 International
  Conference on Management of Data}} (Amsterdam, Netherlands)
  \emph{(\bibinfo{series}{SIGMOD '19})}. \bibinfo{publisher}{Association for
  Computing Machinery}, \bibinfo{address}{New York, NY, USA},
  \bibinfo{pages}{69–86}.
\newblock
\showISBNx{9781450356435}
\urldef\tempurl%
\url{https://doi.org/10.1145/3299869.3319897}
\showDOI{\tempurl}


\bibitem[\protect\citeauthoryear{Van~Laarhoven}{Van~Laarhoven}{2017}]%
        {van2017l2}
\bibfield{author}{\bibinfo{person}{Twan Van~Laarhoven}.}
  \bibinfo{year}{2017}\natexlab{}.
\newblock \showarticletitle{L2 regularization versus batch and weight
  normalization}.
\newblock \bibinfo{journal}{\emph{arXiv preprint arXiv:1706.05350}}
  (\bibinfo{year}{2017}).
\newblock


\bibitem[\protect\citeauthoryear{Wang, Qu, Wu, Wang, and Zhou}{Wang
  et~al\mbox{.}}{2021}]%
        {wang2020we}
\bibfield{author}{\bibinfo{person}{Xiaoying Wang}, \bibinfo{person}{Changbo
  Qu}, \bibinfo{person}{Weiyuan Wu}, \bibinfo{person}{Jiannan Wang}, {and}
  \bibinfo{person}{Qingqing Zhou}.} \bibinfo{year}{2021}\natexlab{}.
\newblock \showarticletitle{Are We Ready for Learned Cardinality Estimation?}
\newblock \bibinfo{journal}{\emph{Proc. VLDB Endow.}} \bibinfo{volume}{14},
  \bibinfo{number}{9} (\bibinfo{date}{May} \bibinfo{year}{2021}),
  \bibinfo{pages}{1640–1654}.
\newblock
\showISSN{2150-8097}
\urldef\tempurl%
\url{https://doi.org/10.14778/3461535.3461552}
\showDOI{\tempurl}


\bibitem[\protect\citeauthoryear{Wu, Ding, Chu, Wei, Dai, Guan, and Zhou}{Wu
  et~al\mbox{.}}{2021}]%
        {url:technical_report}
\bibfield{author}{\bibinfo{person}{Renzhi Wu}, \bibinfo{person}{Bolin Ding},
  \bibinfo{person}{Xu Chu}, \bibinfo{person}{Zhewei Wei},
  \bibinfo{person}{Xiening Dai}, \bibinfo{person}{Tao Guan}, {and}
  \bibinfo{person}{Jingren Zhou}.} \bibinfo{year}{2021}\natexlab{}.
\newblock \bibinfo{title}{{An learned sample-based NDV estimator (technical
  report)}}.
\newblock
  \bibinfo{howpublished}{\url{https://figshare.com/s/8cd5f3dad9418b84b75a}}.
\newblock
\newblock
\shownote{[Online; accessed 11-October-2021].}


\bibitem[\protect\citeauthoryear{Xu, Zhang, Li, Du, Kawarabayashi, and
  Jegelka}{Xu et~al\mbox{.}}{2021}]%
        {xu2020neural}
\bibfield{author}{\bibinfo{person}{Keyulu Xu}, \bibinfo{person}{Mozhi Zhang},
  \bibinfo{person}{Jingling Li}, \bibinfo{person}{Simon~S Du},
  \bibinfo{person}{Ken-ichi Kawarabayashi}, {and} \bibinfo{person}{Stefanie
  Jegelka}.} \bibinfo{year}{2021}\natexlab{}.
\newblock \showarticletitle{How neural networks extrapolate: From feedforward
  to graph neural networks}. In \bibinfo{booktitle}{\emph{ICLR}}.
\newblock


\bibitem[\protect\citeauthoryear{Zhu, Wu, Han, Zeng, Pfadler, Qian, Zhou, and
  Cui}{Zhu et~al\mbox{.}}{2021}]%
        {vldb:ZhuW21}
\bibfield{author}{\bibinfo{person}{Rong Zhu}, \bibinfo{person}{Ziniu Wu},
  \bibinfo{person}{Yuxing Han}, \bibinfo{person}{Kai Zeng},
  \bibinfo{person}{Andreas Pfadler}, \bibinfo{person}{Zhengping Qian},
  \bibinfo{person}{Jingren Zhou}, {and} \bibinfo{person}{Bin Cui}.}
  \bibinfo{year}{2021}\natexlab{}.
\newblock \showarticletitle{FLAT: Fast, Lightweight and Accurate Method for
  Cardinality Estimation}.
\newblock \bibinfo{journal}{\emph{Proc. VLDB Endow.}} \bibinfo{volume}{14},
  \bibinfo{number}{9} (\bibinfo{date}{May} \bibinfo{year}{2021}),
  \bibinfo{pages}{1489–1502}.
\newblock
\showISSN{2150-8097}
\urldef\tempurl%
\url{https://doi.org/10.14778/3461535.3461539}
\showDOI{\tempurl}


\end{thebibliography}
